\pdfoutput=1

\documentclass[11pt]{article}

\usepackage{acl}

\usepackage{times}
\usepackage{latexsym}

\usepackage[T1]{fontenc}

\usepackage[utf8]{inputenc}

\usepackage{microtype}

\usepackage{amsmath,amsfonts,bm}

\def\eqref#1{equation~\ref{#1}}

\def\1{\bm{1}}

\DeclareMathAlphabet{\mathsfit}{\encodingdefault}{\sfdefault}{m}{sl}
\SetMathAlphabet{\mathsfit}{bold}{\encodingdefault}{\sfdefault}{bx}{n}

\usepackage{hyperref}
\usepackage{url}

\setcitestyle{citesep={,}}
\usepackage[utf8]{inputenc} 
\usepackage[T1]{fontenc}    
\usepackage{hyperref}       
\usepackage{url}            
\usepackage{booktabs}       
\usepackage{amsfonts}       
\usepackage{nicefrac}       
\usepackage{microtype}      
\usepackage{xcolor}         
\usepackage{subfiles}
\usepackage{bm}
\usepackage{amstext, amsmath, amsfonts, bbm, amssymb}
\usepackage{graphicx}
\usepackage{caption}
\usepackage{subcaption}
\usepackage{multirow}
\usepackage{wrapfig}
\usepackage{color,soul}
\usepackage{makecell}
\usepackage{cuted,lipsum}

\newcommand{\modelnamett}{\textsc{EISL} }
\newcommand{\modelname}{\textsc{EISL}}

\newcommand{\bmx}{\bm x}
\newcommand{\bmy}{\bm y}

\newcommand{\gram}{\text{gram}}
\newcommand{\precn}{\text{prec}_n}

\usepackage{lipsum}

\title{
Don't Take It Literally: \\An Edit-Invariant 
Sequence Loss for Text Generation
}
\author{Guangyi Liu$^1$,~~
Zichao Yang$^2$,~~
Tianhua Tao$^3$,~~
 Xiaodan Liang$^4$,~~\\
{\bf Junwei Bao$^5$,~~
 Zhen Li$^1$,~~
Xiaodong He$^5$,~~
Shuguang Cui$^1$,~~
Zhiting Hu$^{6}$}\\
$^1$Chinese University of Hong Kong, Shenzhen,~~ $^2$Carnegie Mellon University,~~\\ $^3$Tsinghua University,~~ $^4$Sun Yat-Sen University,~~ $^5$JD AI Research,~~ $^6$UC San Diego\\
{\small \tt guangyiliu@link.cuhk.edu.cn, zhh019@ucsd.edu}
}

\begin{document}

\maketitle
\renewcommand{\thefootnote}{\fnsymbol{footnote}}
\renewcommand{\thefootnote}{\arabic{footnote}}
\begin{abstract}
Neural text generation models are typically trained by maximizing 
log-likelihood with the sequence cross entropy (CE) loss, which encourages an \emph{exact} token-by-token match between
a target sequence with a generated sequence. Such training objective
is sub-optimal when the target sequence is not perfect, e.g.,
when the target sequence is corrupted with noises, or when only weak sequence supervision 
is available. To address the challenge, we propose a novel Edit-Invariant
Sequence Loss (\modelname), which computes the matching loss of a target $n$-gram
with all $n$-grams in the generated sequence. 
\modelnamett is designed to be robust to 
various noises and edits in the target sequences. Moreover, the \modelnamett computation is essentially an approximate convolution operation with target $n$-grams as kernels,
which is easy to implement and efficient to compute with existing libraries. To demonstrate
the effectiveness of \modelname, we conduct experiments on a wide range of tasks, including
machine translation with noisy target sequences, unsupervised text
style transfer with only weak training signals, and non-autoregressive generation with non-predefined generation order. 
Experimental results show our method significantly outperforms
the common CE loss and other strong baselines on all the tasks. EISL has a simple API that can be used as a drop-in replacement of the CE loss.\footnote{Code: \url{https://github.com/guangyliu/EISL}}
\end{abstract}

\section{Introduction}  
\label{sec:intro}

Neural text generation models have ubiquitous applications in natural language processing,
including machine translation~\citep{BahdanauCB14,seq2seq,DBLP:journals/corr/WuSCLNMKCGMKSJL16,transformer}, summarizations~\citep{DBLP:conf/conll/NallapatiZSGX16,DBLP:conf/acl/SeeLM17}, 
dialogue systems~\citep{DBLP:conf/emnlp/LiMRJGG16}, etc.
They are typically 
trained by maximizing the 
log-likelihood of the output sequence conditioning on the inputs with the cross
entropy (CE) loss. The CE loss can be easily factorized into individual loss terms
and can be optimized efficiently with stochastic gradient descent.
Due to its computational efficiency and ease to implement, the training paradigm
has played an important role in building successful large text generation models~\citep{DBLP:journals/corr/abs-1910-13461,radford2019language}.
\begin{figure}
\centering
\includegraphics[width=0.45\textwidth]{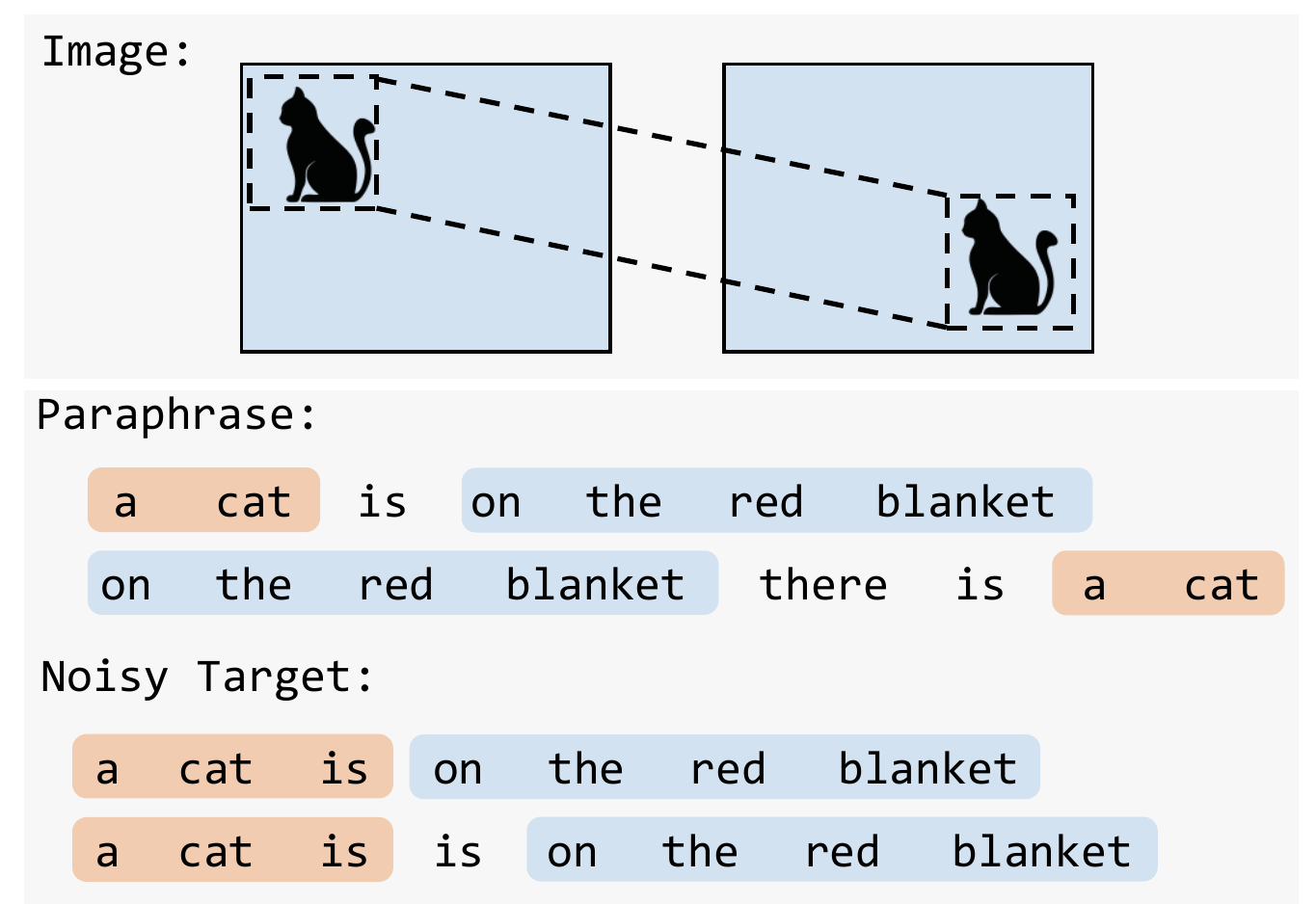}
\vspace{-6pt}
\caption{Invariance exists in both image and text, e.g., image is invariant to translation (top), and text is robust to many forms of edits (bottom).}
\label{fig:introexample}
\vspace{-2pt}
\end{figure}
However, the CE loss minimizes the negative log-likelihood of only the reference output sequence, 
while all other sequences are equally penalized through normalization. This is over-restrictive
since for a given reference target sentence, many possible paraphrases are semantically close,
hence should not completely be treated as negative samples. For example, as shown in Figure~\ref{fig:introexample}, \texttt{a cat is on the red blanket}
should be treated equally with \texttt{on the red blanket there is a cat}. A model trained
with CE loss falls short of modeling such type of invariance for text. 

\begin{figure*}[t]
\centering
\includegraphics[width=0.95\textwidth]{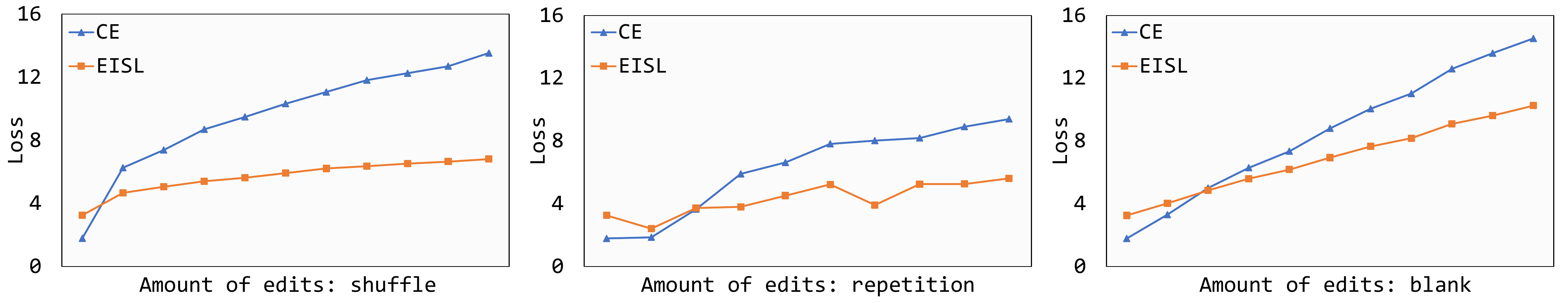}
\vspace{-6pt}
\caption{
Sensitivity of CE and \modelnamett loss w.r.t different types of text edits as the amount of edits increases (x-axis). 
We use a fixed machine translation model, synthesize different types of edits on target text, and measure the CE and \modelnamett losses, respectively. The edit types include shuffle (changing the word order), 
repetition (words being selected are repeated), and word blank (words being replaced with a blank token). 
CE loss tends to increase drastically once a small amount of edits is applied. 
In contrast, \modelnamett loss increases much more slowly, showing its robustness.
}
\label{fig:loss_curve}
\vspace{-6pt}
\end{figure*}

The problem is even exaggerated when the supervision from a target sequence is not perfect \citep{DBLP:conf/wmt/Pinnis18}.
On one hand, there could be {\it noises} in the reference sequence which makes itself not a valid
sentence. As in the last example shown in Figure~\ref{fig:introexample}, 
there is a repetition error in the target sequence, which is common in human generated text. 
With the CE loss, the model is forced to copy all tokens including the error, and
assign a high loss for the grammatically correct sequence. The exact tokens matching renders the CE loss 
sensitive to noises in the target, as shown in Figure~\ref{fig:loss_curve}. On the other hand, there are many problems with only {\it weak} supervision 
for target sequences \citep{tan2020summarizing,wang2021simvlm,lin2020data}. 
For example, in tasks of unsupervised text style transfer \cite{jin2022deep} aiming to rewrite a sentence from one style to another, the original sentence offers weak supervision for the content (rather than the style). Yet using a CE loss here is problematic since it encourages the model to copy every original token. 

Prior works have tried to address this problem using reinforcement learning (RL) \citep{guo2021text,DBLP:journals/corr/abs-1909-03622,wieting-etal-2019-beyond}. 
For example, policy gradient was used to optimize sequence rewards such
as BLEU metric \citep{DBLP:journals/corr/RanzatoCAZ15,DBLP:conf/iccv/LiuZYG017}. Such algorithms assign high rewards
to sentences that are close to the target sentence. Though it is a valid objective to optimize, 
policy optimization faces significant challenges in practice. 
The high variance of gradient estimate makes the training extremely difficult, and 
almost all previous attempts rely on fine-tuning from models trained with CE loss, often with unclear improvement \citep{wu2018study}. 

\begin{figure*}
\centering
\includegraphics[width=0.9\textwidth]{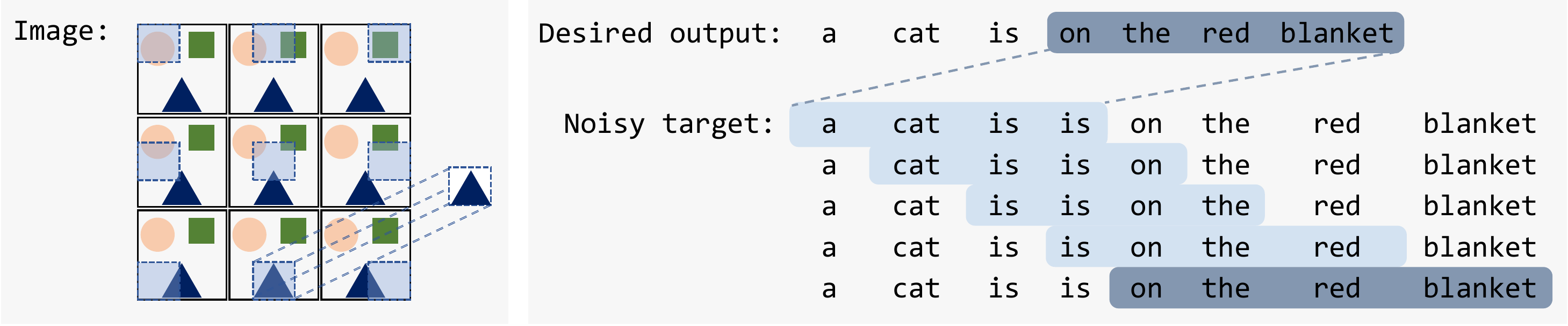}
\vspace{-10pt}
\caption{Inspired by the ConvNet convolution which applies a convolution kernel to different positions in an image and aggregate ({\bf left}), we devise similar $n$-gram matching and convolution, which is robust to sequence edits (noises, shuffle, repetition, etc) ({\bf right}).}
\label{fig:textConv}
\vspace{-15pt}
\end{figure*}

In this paper, we propose an alternative loss to overcome the above weakness of CE loss, 
but reserve all nice properties such as being end-to-end differentiable, easy to implement, and efficient to compute,
which hence can be used as a drop-in replacement or combined with CE. 
The loss is based on the observation that a viable candidate sequence 
shares many sub-sequences with the target. Our loss, called {\it edit-invariant 
sequence loss} (\modelname), models the matching of each reference $n$-gram across all $n$-grams
in a candidate sequence. 
The design is motivated by the translation invariance properties of ConvNets on
images (see Figure~\ref{fig:textConv}), and captures the edit invariance properties of text $n$-grams in calculating the loss. Figure~\ref{fig:loss_curve} shows the invariance property of \modelnamett in comparison with CE.
Appealingly, we show the conventional CE loss is a special case of \modelname---when $n$ equals to the sequence length,
\modelnamett calculates the exact sequence matching loss and reduces to CE. 
Moreover, the computations of \modelnamett is essentially a convolution operation of candidate
sequence using target $n$-grams as kernels, which is very easy to implement with
existing deep learning libraries. 

To demonstrate the effectiveness of \modelnamett loss,  we conduct experiments 
on three representative tasks: machine translation with \emph{noisy} training target, unsupervised
text style transfer (only \emph{weak} references are available), and non-autoregressive generation with \emph{flexible generation order}. Experiments
demonstrate \modelnamett loss can be easily incorporated with a series of sequence
models and outperforms CE and other popular baselines across the board.
\section{Related Work}
Deep neural sequence models such as recurrent neural networks~\citep{seq2seq,rnnlm} and transformers~\citep{transformer} have achieved great progress in many 
text generation tasks like machine translation~\citep{BahdanauCB14,transformer}.
These models are typically trained with the maximum-likelihood 
objective, which can lead to sub-optimal performance due to
CE's exact sequence matching assumption.
%
%
There are lots of works trying to overcome this weakness.
For examples, some works~\citep{DBLP:journals/corr/RanzatoCAZ15, DBLP:conf/cvpr/RennieMMRG17, DBLP:conf/iccv/LiuZYG017, DBLP:conf/acl/ShenCHHWSL16, DBLP:conf/acl/SmithE06a} 
proposed to use policy gradient or minimum risk training to optimize the 
expected BLEU metric \cite{papineni2002bleu}. 
Due to the high variance and unstableness in RL training, a variety of training tricks 
are used in practice. \citet{wieting-etal-2019-beyond} developed a new reward function based
on semantic similarity for translation. \citet{guo2021text} introduced soft Q-learning for more efficient RL training.
On the other hand, \citet{DBLP:journals/corr/abs-1712-04708,DBLP:conf/iclr/CasasFC18} made the initial attempts to develop differentiable BLEU objectives, making soft approximations to the count of $n$-gram matching in the original BLEU formulation.
\citet{DBLP:journals/corr/abs-1809-03132,DBLP:journals/corr/abs-2106-081221,DBLP:conf/aaai/ShaoZFMZ20} minimized the $n$-gram difference between the model outputs and targets in non-autoregressive generation.

Another line of research that is relevant to our work is learning
with noisy labels in classification
\citep{DBLP:conf/nips/ZhangS18,DBLP:conf/nips/XuCKW19,DBLP:conf/iccv/0001MCLY019,hu2019learning}.
For text generation, \citet{DBLP:conf/coling/NicolaiS20} proposed student 
forcing to substitute teacher forcing, which can alleviate the influence of 
noise in the target sequence during decoding.
\citet{DBLP:conf/acl/KangH20} proposed loss truncation, which adaptively removes 
high-loss examples considered as invalid data. Our empirical study shows substantial improvement of our approach over the previous ones.
%

\section{Edit-Invariant Sequence Loss}
\label{sec:eisl}

In this section, we first review the conventional cross-entropy (CE) loss for sequence learning, and point
out its weakness, especially when the target sequence is edited. We then introduce
the \modelnamett loss which gives a model the flexibility to learn from sub-sequences in 
a target sequence.

We first establish notations for the sequence generation setting. 
Let $(\bmx, \bmy^*)$ be a paired data sample where $\bmx$ is the input and $\bmy^*  =(y^*_1,...,y^*_{T^*})$ 
is the reference target sequence. 
Define $\bmy=(y_1,...,y_{T})$ as a candidate sentence.
Our goal is to build a model $p_{\bm\theta}(\bmy|\bmx)$ 
that scores a candidate sequence $\bmy$ with parameter $\bm\theta$.
In the sequel, we omit the condition $\bmx$ and the subscript $\bm \theta$ for simplicity.
\vspace{-0.15cm}
\subsection{The Difficulty of Cross Entropy Loss}
\vspace{-0.1cm}
The standard approach to learn the sequence model is to minimize the negative 
log-likelihood (NLL) of the target sequence, i.e., minimizing the CE loss $\mathcal{L}^\text{CE}(\bm\theta) = -\log p(\bmy^*)$.
The CE loss assumes \emph{exact} matching of a candidate sequence $\bmy$ with the
target sequence $\bmy^*$. In other words, it  maximizes the probability of only the target sequence $\bmy^*$
while penalizing all other possible sequence outputs that might be close but different with $\bmy^*$.

The assumption can be problematic in many practical scenarios: {\bf (1)} For a given target
sentence, there could be many ways of paraphrasing the sentence such as word reordering,
synonyms replacement, active to passive rewriting, etc. Many of the paraphrases are viable 
candidate sequences, and/or share many sub-sequences with the reference sentence, and thus should not be treated completely as negative samples.
Similar to the translation invariance which is shown to be effective in image modeling, a sequence loss that is {\em robust} to the shift and edits of sub-sequences in the reference sequence is preferred in order to model the rich variations of sequences; 
{\bf (2)} The edit-invariance property is particularly desirable when the reference target
sequence is corrupted with noise or is only weak sequence supervision. 
For instance, in Figure~\ref{fig:textConv}, the word \texttt{is} is repeated twice, 
which is one of the common errors in typing. Using CE loss in the noisy target setting forces the model to learn the data errors as well. In contrast, a sequence loss robust or invariant to the shift of sub-sequences assigns a high probability to the correct sentence even though it does not match the noisy target exactly. The loss thus offers flexibility for the model to select right information for learning. 

\vspace{-0.2cm}
\subsection{EISL: Edit-Invariant Sequence Loss}
\vspace{-0.1cm}
Motivated by the above discussion, in this section, we draw inspirations from the convolution operation that enables translation invariance in image modeling (Figure~\ref{fig:textConv}, left), and propose an edit-invariant sequence loss (EISL) as illustrated in Figure~\ref{fig:textConv} (right).
Intuitively, for instance, given a 4-gram \texttt{on the red blanket}, because there is no
extra knowledge to determine the position of the 4-gram in the noisy target sequence, we
compute the losses across all positions in the noisy target sequence and aggregate. This is essentially a convolution
over the target noisy sequence with the given $n$-gram as a convolution kernel. 

We now derive the \modelnamett loss in more details. Let $\bmy_{a:b}=(y_a,...,y_{b-1})$ denote
a sub-sequence of $\bmy$ that starts from index $a$ and ends at index $b-1$, which is of length $b-a$. 
Thus $\bmy^*_{i:i+n}$ denotes the $i$-th $n$-gram in the reference $\bmy^*$. Denote $C(\bmy^*_{i:i+n}, \bmy)$ as the number of times this $n$-gram occurs in $\bmy$:
\begin{equation}
\small
    \label{eq:defCount}
    C(\bmy^*_{i:i+n}, \bmy) = \sum_{i'=1}^{T-n+1}\mathbbm{1}(\bmy_{i':i'+n}=\bmy^*_{i:i+n}),
\end{equation}
where $\mathbbm{1}(\cdot)$ is the indicator function that takes value $1$ if the $n$-grams match, and $0$ otherwise. 
Intuitively, for a text generation model, we would like to maximize 
the occurrence of an $n$-gram from the reference in the target sequence.
For a given probabilistic model $p_{\bm\theta}(\bmy)$ (we omit the parameter $\bm\theta$ wherever
the meaning is clear), the expected value of $C(\bmy^*_{i:i+n},\bmy)$ can be computed as follow:
\begin{equation}
\small
 \begin{split}
&\mathbb{E}_{\bmy \sim p(\bmy)}[C(\bmy^*_{i:i+n},\bmy)]\\
         &= \sum_{i'=1}^{T-n+1}\mathbb{E}_{p(\bmy_{i':i'+n})}\left[\mathbbm{1}(\bmy_{i':i'+n}=\bmy^*_{i:i+n})\right] \\
         &= \sum_{i'=1}^{T-n+1}p(\bmy_{i':i'+n}=\bmy^*_{i:i+n}).
\end{split}
\end{equation}
Thus, for each $i$-th $n$-gram in the reference, a straightforward way to define the learning objective is to minimize the negative log value of its expected occurrence, i.e., $-\log \mathbb{E}_{\bmy \sim p(\bmy)}[C(\bmy^*_{i:i+n},\bmy)]$.

The above loss requires computation of the marginal probability $p(\bmy_{i':i'+n}=\bmy^*_{i:i+n})$ of an $n$-gram,
which is intractable in practice. We therefore derive an upper bound of the loss and use it as the surrogate to minimize in training. We denote the upper bound surrogate as our \modelnamett loss. Specifically, since
for a given $i'$, $p(\bmy_{i':i'+n} = \bmy^*_{i:i+n}) = \sum_{\bmy}p(\bmy_{<i'})p(\bmy_{i':i'+n}=\bmy^*_{i:i+n}|\bmy_{<i'})$,
then:
\begin{equation}
\centering
\small
\begin{split}
    &-\log\mathbb{E}_{\bmy \sim p(\bmy)}[C(\bmy^*_{i:i+n},\bmy)]\\
    &= -\log \sum_{i'=1}^{T-n+1}p(\bmy_{i':i'+n} = \bmy^*_{i:i+n}), \\
    &\leq \frac{-\mathbb{E}_{\bmy \sim p(\bmy)} \sum_{i'=1}^{T-n+1} \log p(\bmy_{i':i'+n}=\bmy^*_{i:i+n} | \bmy_{<i'})}{T-n+1}   \\
    &:= \mathcal{L}^{\text{\modelname}}_{n,i}(\bm\theta). 
\end{split}
\label{eq:eisl-upper-bound}
\end{equation}
The detailed derivation is attached in Appendix~\ref{app:derive}. Notice that the \modelnamett loss involves only the conditional distribution $p(\bmy_{i':i'+n}=\bmy^*_{i:i+n} | \bmy_{<i'})$ which is convenient to compute---we first sample tokens from the model up to the $i'$ position, then compute NLL of the reference $n$-gram $\bmy^*_{i:i+n}$ occurring at position $i'$ under the model distribution.
The full $n$-gram EISL loss is then defined by averaging across all $n$-gram positions in the reference:
\begin{equation}
\small
    \label{eq:eisln}
      \mathcal{L}^{\modelname}_n(\bm \theta) = \frac{1}{T^*-n+1}\sum_{i=1}^{T^*-n+1}\mathcal{L}_{n,i}^{\modelname}(\bm \theta).
\end{equation}
In practice, inspired by the standard BLEU metric (more in section~\ref{sec:connections}), we could also straightforwardly combine different $n$-gram losses depending on tasks:
\begin{equation}
\small
    \mathcal{L}^{\modelname}(\bm\theta) = \sum\nolimits_{n} w_n \cdot\mathcal{L}^{\modelname}_{n}(\bm\theta),
\end{equation}
where $w_n$ is the weight of the $n$-gram loss.
The rule of thumb is that a $n$-gram EISL loss with lower $n$ is more robust to noises, as shown in our experiments.
Following BLEU, we found that simply using equal weights for different $n$-grams up to $n=4$ often produces good performance.

As discussed shortly, it is appealing that the $n$-gram \modelnamett loss is indeed a direct generalization of the CE loss on the $n$-gram
level: we sum the CE loss of an $n$-gram over all candidate sequence positions by conditioning
on samples from the model.
Besides, the derivation of the upper bound makes no assumption on the probability function $p(\bmy)$, hence holds for both autogressive and non-autoregressive sequence models as demonstrated in our experiments.
\begin{figure*}
\centering
\includegraphics[width=0.85\textwidth,page=1]{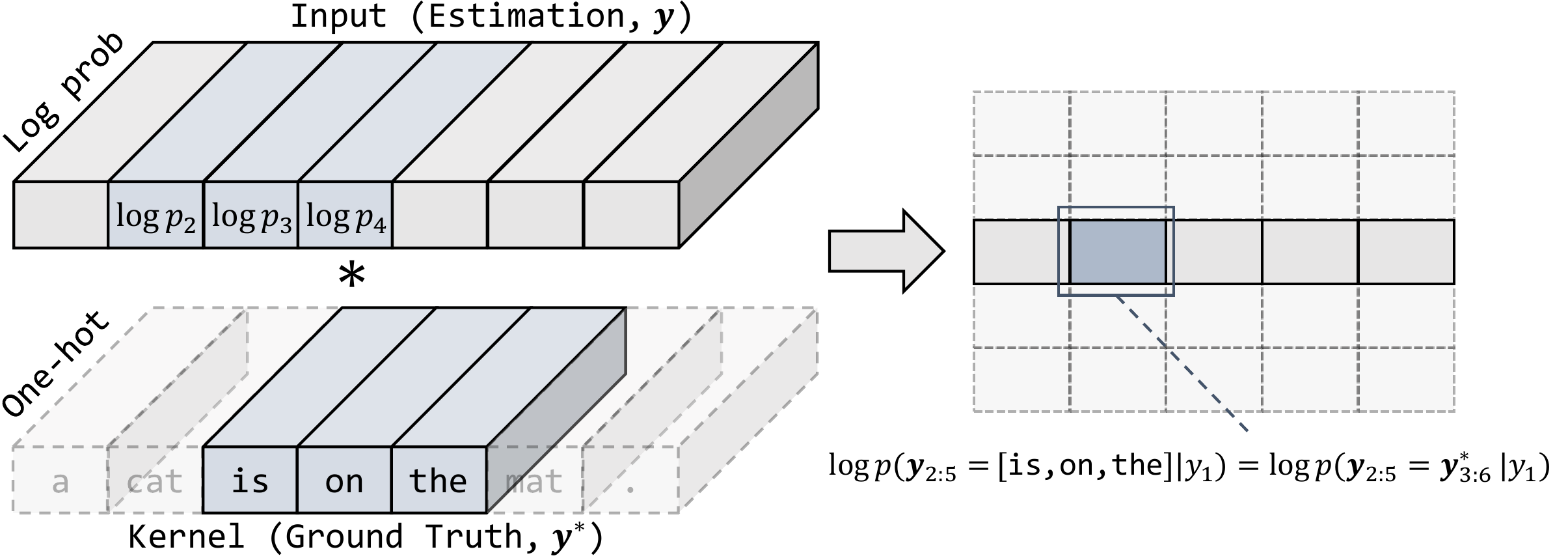}
\vspace{-6pt}
\caption{As convolution is a common operation for translation invariance in 
image, we adopt a convolution to achieve the translation invariance in text. 
The input is the distribution from the model output in log domain, 
kernel represents the convolution kernel and $*$ is the convolution operation.
In this $3$-gram example, there are 5 kernels, which correspond to the 5
rows on the right. }
\label{fig:conv}
\vspace{-6pt}
\centering
\end{figure*}

\textbf{Position Selection} 
Minimizing the gram matching loss over all positions can make the model assign equal probabilities at all positions, which causes the training to collapse. 
We further adapt the loss to enable the model to
automatically learn the positions of reference $n$-grams. 
For notation simplicity, let $g_{i,i'}^n$ denote the conditional probability $p(\bmy_{i':i'+n}=\bmy^*_{i:i+n}|\bmy_{<i'})$ involved above (Eq.\ref{eq:eisl-upper-bound}).
We can vectorize the probability to get $\bm g_i^n = [g_{i,1}^n,...,g_{i,T-n+1}^n]^T$, spanning all potential positions in the candidate sequence.
We then normalize the probability vector $\bm g_i^n$ by Gumbel softmax \citep{JangGP17}, denoted as 
$\bm q_i^n = \texttt{Gumbel\_softmax}(\bm{g}_{i}^n)$, which we use as the weight for
every $n$-gram positions.
We multiply the weight with the original log probability
to get the new adjusted loss:
\begin{equation}
\small
\label{eq:eisl_loss}
    \mathcal{L}_{n,i}^{\text{\modelname}} (\bm\theta)
    \approx -\bm q_{i}^n \cdot \log\bm{g}_{i}^n.
\end{equation}
The loss can roughly be viewed as the ``entropy'' of the unnormalized probabilities $\bm{g}_i^n$, which has minimal value if the mass of the probability is assigned to one location only. 
Intuitively, if an $g_{i,i'}^n$ is large, then it is likely $i'$ is
the correct position for the reference $n$-gram, hence the weight for this position should also be large. This is like the 
greedy exploitation in reinforcement learning~\citep{MnihKSRVBGRFOPB15}. 
On the other hand, to overcome over-exploitation, the Gumbel softmax introduces randomness in the weight assignment, which helps balance the exploitation-exploration trade-off in position selection for the model.

\label{para:approx_p}
\textbf{Efficient Approximate Computation: \modelnamett as Convolution} We show the \modelnamett loss can be computed efficiently using the common convolution operator, with very little additional cost compared with the CE loss.  The computation involves moderate approximation if the generation model is an autoregressive model, and is exact in the case of a non-autoregressive model (e.g., as in section~\ref{sec:exp:nat}). 
We first discuss the easy case when the model is a non-autoregressive model, where we have $g_{i,i'}^n = p(\bmy_{i':i'+n}=\bmy^*_{i:i+n}|\bmy_{<i'}) = \prod_{j=1}^{n} p(y_{i'+j-1}=y^*_{i+j-1})$.
Denote $V$ as the vocabulary size. Let $\bm P = [\bm p_1, \bm p_2, ... \bm p_T]$ be the probability output by the model across positions, where $\bm p_{i'} \in \mathbb{R}^{V}$ is the probability output after softmax at $i'$-th position, and each $\bm p_{i'}$ is independent with each other. 
On this basis, we compute the key quantity $\log \bm g_i^n$ in Eq.~\ref{eq:eisl_loss} as the direct output of the convolution operator. 
As shown in Figure~\ref{fig:conv}, we can get $\log \bm g_i^n$ by applying
convolution on $\log \bm P$, with $\bm y_{i:i+n}$ as the kernels:
\begin{equation}
\small
    \label{eq:conv}
     \log \bm g_i^n= \texttt{Conv}(\log\bm P, \texttt{Onehot}(\bmy^*_{i:i+n})),
\end{equation}
where $\texttt{Onehot}(\cdot)$ maps each token to its corresponding one-hot representation
and $\texttt{Conv}(\cdot,\cdot)$ is the convolution operation with the first argument as input
and the second as the kernel. We transform $\bm P$ into log domain to turn the probability multiplication into log probability summations, where $\texttt{Conv}$ 
can be directly applied. 
As shown in Figure~\ref{fig:conv}, $\log\bm P$
is of shape $V \times T$ and $\texttt{Onehot}(\bm y^*_{i:i+n})$ is of shape $V \times n$,
so $\texttt{Conv}(\log\bm P, \texttt{Onehot}(\bmy^*_{i:i+n}))$ is an one-dimensional convolution on the sequence axis.  
Formally, the $i'$-th convolutional output is:
\begin{equation}
\small
\begin{split}
        \log g_{i,i'}^n 
        =& \sum_{j=1}^{n} \log\bm p_{i'+j-1}\cdot \texttt{Onehot}(y^*_{i+j-1}) \\
        =& \sum_{j=1}^{n}\log p(y_{i'+j-1}=y^*_{i+j-1}|\bmy_{<i'+j-1})
\end{split}
\end{equation}

After obtaining $\bm g_{i}^n$ by convolution, the \modelnamett loss in Eq.~\ref{eq:eisl_loss} can be easily calculated.
\begin{figure*}
\centering
\includegraphics[width=0.95\textwidth]{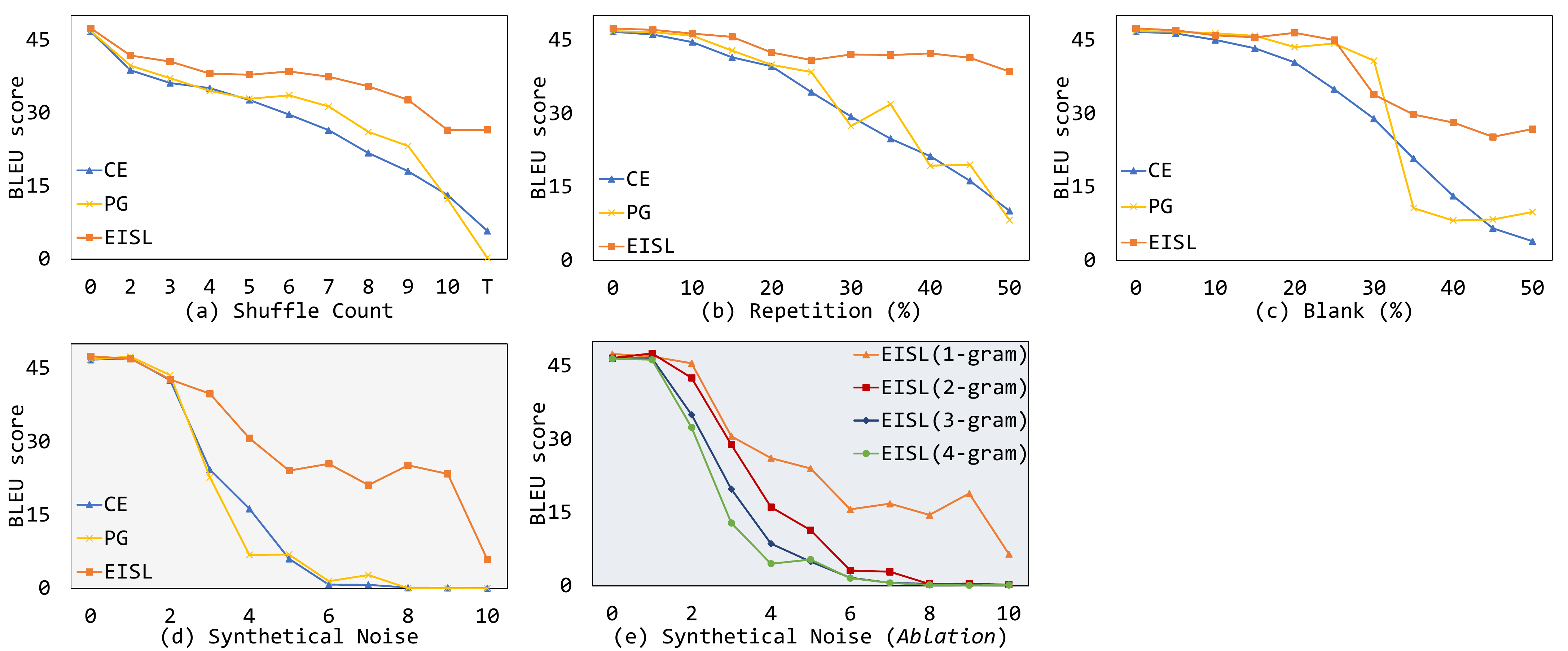}
\caption{Results of Translation with Noisy Target on German-to-English(de-en) from Multi30k. BLEU scores are computed 
against clean test data. The $x$-axis of all figures denotes the level of noise we 
injected to target sequences in training.
(a) Shuffle: selected tokens are shuffled;
(b) Repetition: selected tokens are repeated; 
(c) Blank: selected tokens are substituted with a special blank token; 
(d) Synthetical noise: the combination of all three noises ($x=x_0$ stands for the combination of $5x_0\%$ of all kinds of noises);
(e) Ablation study of $n$-grams for \modelnamett on synthetical noise.
BLEURT results are shown in Appendix~\ref{app:noisy_results}.
}
\label{fig:exp1result}
\vspace{-15pt}
\end{figure*}
 We now discuss the case of autoregressive model, where by definition we have $g_{i,i'}^n = \prod_{j=1}^{n} p(y_{i'+j-1}=y^*_{i+j-1} | \bmy_{<i'}, \bmy^*_{i:i+j-1} )$. The dependence on both $\bmy_{<i'}$ and $\bmy^*_{i:i+j-1}$ in each conditional makes exact estimation of $\log \bm g_i^n$ very complicated and costly. We thus introduce the approximation where we approximate $g_{i,i'}^n$ as $\widetilde{g}_{i,i'}^n = \prod_{j=1}^{n} p(y_{i'+j-1}=y^*_{i+j-1} | \bmy_{<i'+j-1} )$. That is, instead of conditioning on $\bmy^*_{i:i+j-1}$, we use the model-generated tokens $\bmy_{i':i'+j-1}$ as the condition. This simple approximation enables us to define the probability output $\bm P$ as in the non-autoregressive case, by just performing a forward pass of the model (i.e., sampling a token $\bmy_i'$ for each position $i'$ and feeding it to the next step to get $\bm p_{i'+1}$). We can then apply the same convolution operator to approximately obtain $\log \bm g_i^n$ as in Eq.~\ref{eq:conv}. Besides the great gain of computational efficiency, we note that the approximation is also effective, especially due to the \emph{position selection} discussed above. Specifically, for each reference $n$-gram $\bmy^*_{i:i+n}$, the position selection in effect (softly) picks those large-value $g_{i,i'}^n$ (while dropping other low-value ones) to evaluate the loss. A large $g_{i,i'}^n$ value indicates the candidate $\bmy_{i':i'+n}$ is highly likely to match the reference $\bmy^*_{i:i+n}$, meaning that using $\bmy_{i':i'+n}$ in replacement of $\bmy^*_{i:i+n}$ is a reasonable approximation for evaluating the above conditionals. We provide empirical analysis of the approximation in Appendix~\ref{sec:app:approximation}, where we show the efficient approximate EISL loss values are very close to the exact EISL values.

\subsection{Connections with Common Techniques}\label{sec:connections}
%
\paragraph{CE is a special case of \modelname}
A nice property of \modelnamett is that it subsumes the standard CE loss as a special case. To see this, set $n=T^*$ (the target sequence length), and we have:
\begin{equation*}
\small
\begin{split}
    \mathcal{L}^{\modelname}_{T^*} = \mathcal{L}_{T^*,1}^{\modelname} 
    = -\log{\bm g}^{T^*}_{1}=-\log p(\bmy=\bmy^*)=\mathcal{L}^{\text{CE}}.
\end{split}
\end{equation*}
The connection shows the generality of \modelname. As a generalization of CE, it enables learning at arbitrary $n$-gram granularity.
\paragraph{Connections between BLEU and \modelname}
Both our method and the popular BLEU~\citep{bleu_paper} metric use $n$-grams as
the basis in formulation. Here we articulate the connections and difference between the two.
Let us first take a review of the BLEU metric.
Specifically, BLEU is defined as a weighted geometric mean of $n$-gram precisions:
\begin{equation*}
\small
    \begin{split}
    \text{BLEU} &= \text{BP}\cdot\exp\left(\sum_{n=1}^Nw_n\log\text{prec}_n\right)\\
    \precn &= \frac{ \sum_{s\in\gram_n(\bmy)}\min(C(\bm s,\bmy),C(\bm s,\bmy^*)) }{ \sum_{s\in\gram_n(\bmy)} C(\bm s,\bmy)},
    \end{split}
\end{equation*}
where BP is a brevity penalty depending on the lengths of $\bmy$ and $\bmy^*$; $N$ is the maximum $n$-gram order (typically $N=4$); \{$w_n$\} are the weights which usually take $1/N$; $\text{prec}_n$ is the $n$-gram precision, $\gram_n(\bmy)$ is the set of unique $n$-gram sub-sequences of $\bmy$; and $C(\bm s,\bmy)$ is the number of times a gram $s$ occurs in $\bmy$ as defined in Eq.~\ref{eq:defCount}.
The conventional formulation above enumerates over unique $n$-grams in $\bmy$. 
In contrast, we enumerate over token indexes in calculating the $n$-gram matching loss.
BLEU considers the $n$-gram precisions and has a penalty term while \modelnamett
simply maximizes the log probability of $n$-gram matchings.

The non-differentiability of BLEU makes it hard to optimize directly, hence
most prior attempts resort to reinforcement learning algorithms and use BLEU
as the reward~\citep{DBLP:journals/corr/RanzatoCAZ15,DBLP:conf/iccv/LiuZYG017}. There are also some works trying to introduce
differentiable BLEU metric using approximation like \cite{DBLP:journals/corr/abs-1712-04708}.
However, such losses are often too complicated and have not yet demonstrated
to perform well in practice.

\section{Experiments}
\label{sec:exp}
In this section, we present the experimental results on three text generation settings to test \modelname's effectiveness, including
learning from noisy text, learning from weak sequence supervision, and non-autoregressive
generation models that require flexibility in generation orders. 
{More details of the experimental setting  are provided in Appendix~\ref{app:setting}.}


\subsection{Learning from Noisy Text}
\label{sec:noisy}
To test the robustness to noise, we evaluate on the task of machine translation with noisy training target, 
in which we train the models with noisy sequence targets and evaluate with clean test data.


\textbf{Setup}
We test \modelnamett loss on Multi30k and WMT18 raw corpus. 
We use German-to-English (de-en) dataset from Multi30k~\citep{DBLP:journals/corr/ElliottFSS16}, which contains 29k training instances.
As inspired by \citet{DBLP:journals/corr/abs-1905-12777}, to simulate various noises in the real data, 
we introduce four types of noises: shuffle, repetition, blank, and the synthetical noise, 
i.e., the combination of the aforementioned three types of noise.
The noises are only added to the training target sequences.
To verify the validity of EISL on real noisy data, we also use German-to-English (de-en) dataset 
from \href{http://www.statmt.org/wmt18/parallel-corpus-filtering.html}{WMT18 raw corpus},
which is a very noisy de-en corpus crawled from the web. 
We randomly select different number of training samples to test the influence of the data scale.
We use a Transformer-based pretrained model BART-base~\citep{DBLP:journals/corr/abs-1910-13461}
 and adopt greedy decoding in training and beam search 
(beam size $=5$) in evaluation. 
We compare \modelnamett loss with CE loss, Policy Gradient (PG), and Loss Truncation (LT). 
We also conduct ablation experiments to explore the effect of different $n$-grams
in \modelnamett loss.
{
We use both BLEU~\citep{bleu_paper} and BLEURT, an advanced model-based metric \citep{DBLP:journals/corr/abs-2004-04696}, as the automatic metrics for evaluation.} Due to space limit, we report BLEU results in the main paper, and defer BLEURT results in the appendix, where we can see BLEURT leads to the same conclusion as BLEU.

\begin{figure}
\centering
\includegraphics[width=0.45\textwidth]{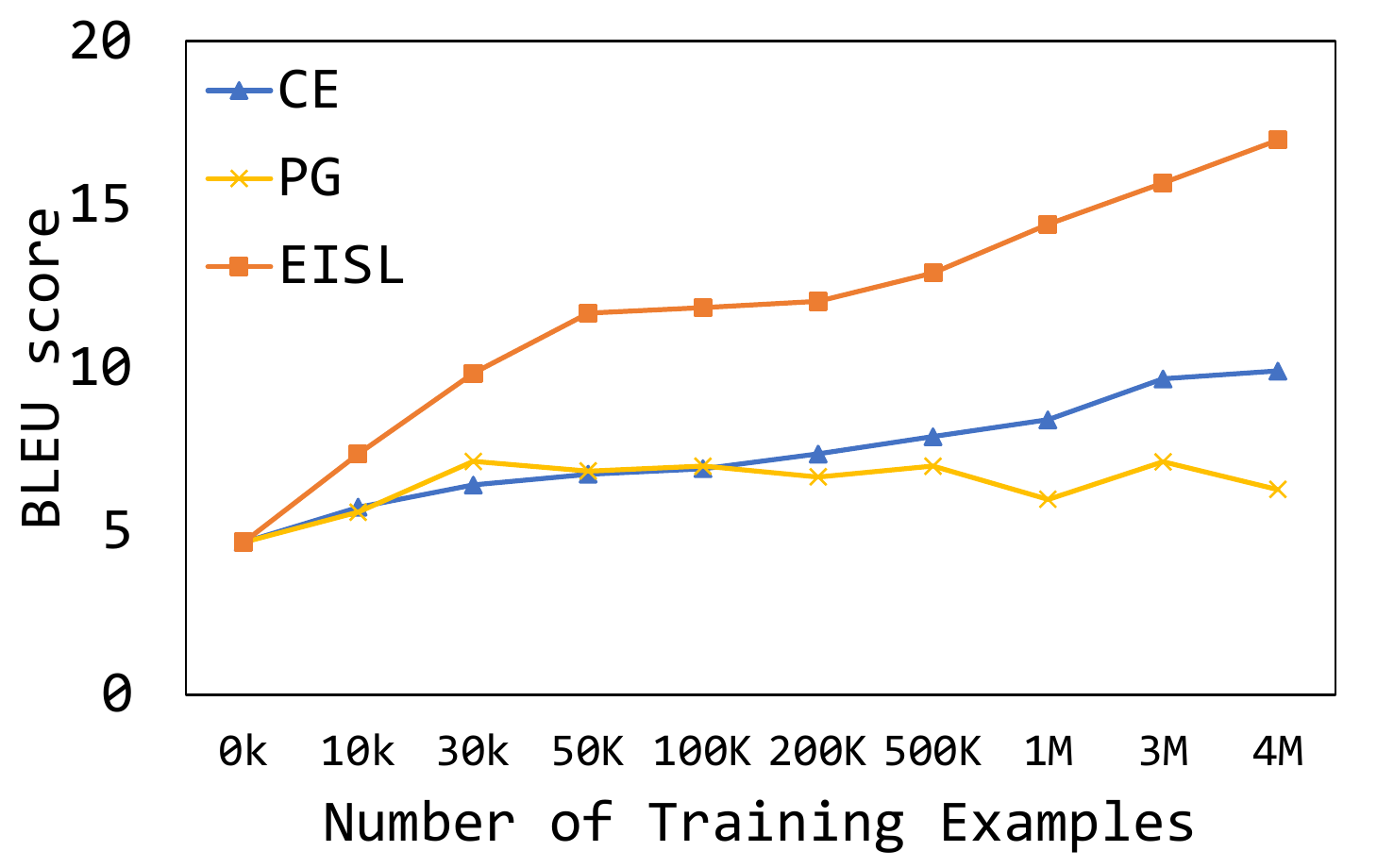}
\vspace{-5pt}
\caption{Results of German-to-English(de-en) Translation on WMT18 raw corpus. BLEU scores are computed against clean parallel test data. 
On x-axis, 0k denotes the performance of the pretrained model. BLEURT results are similar as shown in Appendix~\ref{app:noisy_results}.}
\label{fig:wmt_raw}
\vspace{-10pt}
\end{figure}
\begin{table}[t]
\scriptsize
\begin{center}
\begin{tabular}{@{}rccccc@{}}
    \cmidrule[\heavyrulewidth]{1-6}
    \multirow{2}{*}{\bf Model}&\bf Acc&\multirow{2}{*}{\bf BLEU}& \bf BLEU &\multirow{2}{*}{\bf PPL}& \bf POS\\
    & \bf (\%) & & \bf (Human) & & \bf Distance\\
    \cmidrule{1-6}
    \citet{DBLP:conf/icml/HuYLSX17} &86.7&58.4&-&177.7&-\\
    \citet{DBLP:conf/nips/ShenLBJ17}&73.9&20.7&7.8&72.0&-\\
    \citet{DBLP:conf/iclr/HeWNB20}&87.9&48.4&18.7&\bf31.7&-\\
    \citet{DBLP:journals/corr/abs-1905-05621}&87.7 & 54.9 & 20.3 & 73.0& -\\
    \cmidrule{1-6}
    \citet{DBLP:journals/corr/abs-1810-06526} &\bf88.8 & 65.71 & 22.56 & 42.07& 0.352\\
    \textit{with} \modelname~(Ours)  &\bf88.8 &\bf{68.51} &\bf23.17 &41.56 & \bf0.275 \\ 
    \cmidrule[\heavyrulewidth]{1-6}
\end{tabular}
\begin{tabular}{lll}
    \cmidrule[\heavyrulewidth]{1-3}
    \multicolumn{1}{l}{\bf \citet{DBLP:journals/corr/abs-1810-06526}~(\%)}  &\multicolumn{1}{l}{\bf \textit{with} \modelname~(Ours)~(\%)} & \multicolumn{1}{l}{\bf equal~(\%)} \\ 
    \cmidrule{1-3}
    22.0&30.7 & 47.3\\
    \cmidrule[\heavyrulewidth]{1-3}
\end{tabular}
\vspace{-10pt}
\caption{{\bf Top:} automatic evaluations on the Yelp review dataset.
The BLEU (human) is calculated using the 1000 human annotated sentences as ground truth from \citet{li-etal-2018-delete}. 
The first four results are from the original papers.
{\bf Bottom:} human evaluation statistics of base model vs.\ \textit{with} \modelname.
The results denotes the percentages of inputs for which the model has better
transferred sentences than other model.}
\label{tab:yelp_result}
\end{center}
\vspace{-20pt}
\end{table}
\textbf{Results} 
The results on noisy Multi30k are presented in Figure~\ref{fig:exp1result}. 
The proposed \modelnamett loss provides significantly better performance than 
CE loss and PG on all the noise types, 
especially on the high-level noise end. For synthetical noise as shown in 
Figure~\ref{fig:exp1result}(d), it's interesting to see that CE and PG completely 
fail when the noise level is beyond $6$, but model trained with \modelnamett has
high BLEU score, demonstrating \modelnamett can select useful information to learn despite high noise.
This validates that the proposed \modelnamett is much less sensitive to the noise
than the traditional CE loss and policy gradient training method. 
The results of different $n$-gram are shown in Figure~\ref{fig:exp1result}(e). As the noise increases, the importance of lower grams, e.g., $1$-gram, is more obvious. 
The results on real noisy data, WMT18 raw data, are shown in Figure~\ref{fig:wmt_raw}. 
\modelnamett loss achieves better performance than CE loss and PG, and the difference 
is getting larger when the training data scale increases. 
This again demonstrates \modelnamett could learn more valid information in rather noisy data, while CE loss which only considers whole-sentence matching 
could struggle on noisy data. In Appendix~\ref{app:noisy_results}, we provide more results (e.g., comparison with loss truncation~\citep{DBLP:conf/acl/KangH20}) and case studies.


\begin{table*}[htb]
\footnotesize
  \centering
  \begin{tabular}{llcccc}
    \cmidrule[\heavyrulewidth]{1-6}
     \multirow{2}{*}{\bf Decoding method}   &   \multirow{2}{*}{\bf Model}  
     &  \multicolumn{2}{c}{\bf WMT14 en-de KD}   &   \multicolumn{2}{c}{\bf WMT14 en-de} \\
    \cmidrule(lr){3-4}\cmidrule(lr){5-6} 
    &  & CE & \modelname & CE & \modelname \\
    \midrule
    Autoregressive & Transformer base~\citep{transformer} & \multicolumn{4}{c}{27.48}     \\
    \midrule
    \multirow{5}{*}{Non-Autoregressive}&Vanilla-NAT~\citep{natbase}  &   17.9    &   {\bf22.2} &   9.12    & {\bf15.46}   \\
    & NAT-CRF~\citep{DCRF} & 21.88   &  {\bf22.43}  & -& - \\
    &iNAT~\citep{inat}   & 16.67 & \bf 22.59 & - & - \\
    &LevT~\citep{levt}  & 17.84 & \bf 23.61 & 9.91 & \bf 18.47 \\
    &CMLM~\citep{cmlm} & 17.12 & \bf 23.05 & - &  - \\
    \cmidrule[\heavyrulewidth]{1-6}
  \end{tabular}
  \vspace{-10pt}
    \caption{
    The test-set BLEU of \modelnamett loss and CE loss applied to non-autoregressive models. ``KD'' refers to the standard ``knowledge distillation'' setting in NAT~\citep{natbase}. 
    iNAT, LevT and CMLM are iterative non-autoregressive models,
    that could run in multiple decoding iterations. 
    However, the first decoding iteration of these models is fully
    non-autoregressive, which is what we use as our baselines.}
    \label{table:NAT-table}
  \begin{tabular}{lc}
    \cmidrule[\heavyrulewidth]{1-2}
     {\bf Fully Non-Autoregressive model}  & {\bf WMT14 en-de KD}\\
      \midrule
     \quad CMLM \textit{with} CE~\citep{cmlm} &  17.12\\
    \quad Auxiliary Regularization~\citep{aux_reg}& 20.65\\
    \quad Bag-of-ngrams Loss~\citep{DBLP:conf/aaai/ShaoZFMZ20}        &  20.90 \\
    \quad Hint-based Training~\citep{hint_training}     & 21.11 \\
    \quad CMLM \textit{with} AXE~\citep{axe}    & 23.53\\
    \quad CMLM \textit{with} EISL ({\bf Ours}) & {\bf 24.17}\\
    \midrule
    \cmidrule[\heavyrulewidth]{1-2}
  \end{tabular}
  \vspace{-10pt}
    \caption{
    The test-set BLEU of CMLM trained with our \modelname, compared to other recent fully non-autoregressive methods. The baseline results are from \cite{axe}, where CMLM-with-AXE generates 5 candidates and ranks with loss. Our method follows the same generation configuration as CMLM-with-AXE.
    }
    \label{table:NAT-baselines}
\vspace{-15pt}
\end{table*}
\subsection{Learning from Weak Supervisions: Style Transfer}
We experiment on transferring two types of text styles \cite{jin2022deep}, namely sentiment and political slant, 
to verify \modelnamett can learn from weak sequence supervisions. 

\textbf{Setup}
We use the Yelp review dataset and political dataset. Yelp contains almost 250k negative sentences and 380K positive sentences, of which the ratio of training, valid and test is $7:1:2$. \citet{li-etal-2018-delete} annotated 1000 sentences as ground truth for better evaluation.
The political dataset is comprised of top-level comments on Facebook posts from all 412 members of the United States Senate and House who have public Facebook pages~\citep{voigt-etal-2018-rtgender}. 
The data set contains 270K democratic sentences and 270K republican sentences. And there exists no ground truth for evaluation.
The data preprocessing follows \citet{DBLP:journals/corr/abs-1810-06526}.
The structured content preserving model \citep{DBLP:journals/corr/abs-1810-06526} is adopted as the base model.

Following previous work, we compute automatic evaluation metrics: accuracy, BLEU score, perplexity (PPL) and POS distance.
We also perform human evaluations on Yelp data to further test the transfer quality. 

\textbf{Results}
As sentiment results are shown in Table~\ref{tab:yelp_result}, the BLEU gets improved from 65.71 to 68.51 with \modelnamett loss. On the premise of the correctness of sentiment transfer, \modelnamett loss plays a critical role to guarantee lexical preservation. In the meanwhile, all of BLEU(human), PPL, and POS distance get improved.  It is not surprising that \modelnamett loss helps generate sentences more fluently and select the more appropriate words conditions on the content information. 
As the human evaluation results are shown in Table~\ref{tab:yelp_result}, the model with \modelnamett loss performs better, in accord with the automatic metrics. After analyzing the generated samples, we found \modelnamett loss could drive the model to adopt the words which fit the scene better and could understand more semantics but not just replace some keywords. 
See some examples in the Appendix~\ref{app:tst_yelp}.

We report the results of political data in Appendix~\ref{app:tst_pol}.
Our method outperforms all models on BLEU, PPL, and POS distance with comparable accuracy. 
For a more fair comparison with the base model, our \modelnamett loss improves the base model
on all four metrics, including the accuracy.

The results demonstrate the effectiveness of \modelnamett for weak supervision task, improving both transfer accuracy fluency and content preservation.

\subsection{Learning Non-Autoregressive Generation}\label{sec:exp:nat}
Non-autoregressive neural machine translation (NAT, \cite{natbase}) is proposed 
to predict tokens simultaneously in a single decoding step, which aims at reducing the inference latency. 
The non-autoregressive nature makes it extremely hard for models to keep the order
of words in the sentences, hence CE often struggles with NAT problems.
In experiments, we show EISL is superior to CE in NAT which
requires modeling flexible generation order of the text.

\textbf{Setup}
We use English-to-German dataset from WMT14~\citep{luong-pham-manning}, which contains 4.5M training instances.
We apply our proposed \modelnamett loss on both fully NAT models \citep{natbase,DCRF} and iterative NAT models \citep{inat,levt,cmlm}, showing its general applicability and superiority, and we also compare with a wide range of recent methods~\citep{DBLP:conf/aaai/ShaoZFMZ20,aux_reg,hint_training,axe}. 
We evaluate with both BLEU and BLEURT metrics.

\textbf{Results} 
We first summarize the comparison of BLEU between \modelnamett loss and CE loss in Table~\ref{table:NAT-table} (comparison of BLEURT is in Appendix~\ref{app:nat_bleurt}).
The proposed \modelnamett improves the model performance on both the KD and original datasets. 
More specifically, for fully NAT models (Vanilla-NAT and NAT-CRF), \modelnamett gives strong improvement. 
For iterative NAT models (iNAT, LevT, and CMLM),
\modelnamett also significantly outperforms the baselines when the iteration step is restricted to a small level as suggested by \citet{Kasai2020DeepES}.
(We show in Appendix~\ref{app:nat_iter} that, with increasing iteration steps, 
the difference fades away.
However, as studied in \citet{Kasai2020DeepES}, iterative NAT models with many iteration steps do not hold the intrinsic advantage of speed since Transformer baselines with a shallow decoder can achieve comparable speedup and only at the sacrifice of minor performance drop.) 
Table~\ref{table:NAT-baselines} provides more comparison of with recent strong baselines. Specifically, we apply our EISL on the CMLM base model \cite{cmlm} which shows strong superiority.
We provide qualitative analysis in Appendix~\ref{app:nat_qual}.

\section{Conclusions}
\label{sec:conclusion}
We have developed Edit-Invariant Sequence Loss (\modelname) for end-to-end 
training of neural text generation models. 
The proposed method is insensitive to the shift of $n$-grams in target sequences,
hence suitable for training with noisy data and weak supervisions,  where CE loss fails easily.
We show CE loss is a special case of \modelnamett and build 
the connection of EISL with BLEU metric and convolution operation, 
which both have the invariant property. Experiments on translation with noisy target, 
text style transfer, and non-autoregressive neural machine translation demonstrate the
superiority of our method. 
The more general applications and superiority of EISL on other diverse text generation problems as well as fundamental challenges, such as compositional generalization \cite{andreas2019measuring} and causal invariance \cite{hu2021causal} in language, remain to be explored further, which we are excited to study in the future.



\bibliographystyle{acl_natbib}
\bibliography{naacl_bib}
\appendix
\newpage
\section{Appendix}
\subsection{Additional Derivation}
\label{app:derive}
For a given $i'$,
\begin{align*}
    p(&\bmy_{i':i'+n} = \bmy^*_{i:i+n})\\ 
    &=\sum_{\bmy}p(\bmy_{<i'})p(\bmy_{i':i'+n}=\bmy^*_{i:i+n}|\bmy_{<i'}),
\end{align*}
then we derive the detail of Eq.~\ref{eq:eisl-upper-bound} in Eq.~\ref{eq:add_derive}, where the first inequality holds since $T-n+1\geq 0$; and the second inequality holds by Jensen's inequality.
\begin{figure*}[t] 
\begin{align}
\label{eq:add_derive}
    l^{\text{\modelname}}_{n,i}(\bm\theta)
    &= -\log \sum_{i'=1}^{T-n+1}p(\bmy_{i':i'+n} = \bmy^*_{i:i+n}), \\
    &= -\log \frac{1}{T-n+1}\sum_{i'=1}^{T-n+1}\sum_{\bmy}p(\bmy_{<i'})p(\bmy_{i':i'+n}=\bmy^*_{i:i+n} | \bmy_{<i'}) - \log(T-n+1), \nonumber \\
    &\leq -\log \frac{1}{T-n+1}\sum_{i'=1}^{T-n+1}\sum_{\bmy}p(\bmy_{<i'})p(\bmy_{i':i'+n}=\bmy^*_{i:i+n} | \bmy_{<i'}), \nonumber \\
    &\leq -\frac{1}{T-n+1}\sum_{i'=1}^{T-n+1}\sum_{\bmy} p(\bmy_{<i'}) \log p(\bmy_{i':i'+n}=\bmy^*_{i:i+n} | \bmy_{<i'}), \nonumber \\
    &= -\frac{1}{T-n+1} \mathbb{E}_{\bmy \sim p(\bmy)} \sum_{i'=1}^{T-n+1} \log p(\bmy_{i':i'+n}=\bmy^*_{i:i+n} | \bmy_{<i'}), \nonumber\\
    & = \mathcal{L}^{\text{\modelname}}_{n,i}(\bm\theta), \nonumber
\end{align}
\end{figure*}
\subsection{Detailed Experimental Setup}\label{app:setting}
\subsubsection{Learning from Noisy Text}\label{app:noisy_set}
We use a Transformer-based pretrained model BART-base~\citep{DBLP:journals/corr/abs-1910-13461}, 
containing 6 layers in the encoder and decoder.
We train the model using the Adam optimizer with learning rate $3\times10^{-5}$ with polynomial 
decay and the maximum number of tokens is 6000 in one step. 
The models are trained on one Tesla V100 DGXS with 32GB memory.
We start with CE training using teacher forcing for fast initialization.
We then switch to combined $1$- and $2$-gram \modelnamett with weight $0.8:0.2$,
which we select using the validation set.
We adopt greedy decoding in training and beam search 
(beam size $=5$) in evaluation. 
We use fairseq\footnote{Fairseq(-py) is MIT-licensed.}~\citep{fairseq} to conduct the experiments. 
We compare \modelnamett loss with CE loss and Policy Gradient (PG), 
where PG is used to finetune the best CE model. Teacher forcing is employed in CE training.
\subsubsection{Learning from Weak Supervisions: Style Transfer}\label{app:style_set}
We use the Adam optimizer with learning rate $5\times10^{-4}$, the batch size is $128$ and the model is trained on one Tesla V100 DGXS 32GB.
We compare the results between the base model and the model with \modelname. Specifically, 
on top of the base model, we add the \modelnamett loss (a combination of $2$, $3$ and $4$-gram 
with the same weights $1/3$) to reduce the discrepancy between the transferred sentence 
generated by language model and the original sentence. 
We assign \modelnamett loss with weight $0.5$.

Following previous work, we compute automatic evaluation metrics: accuracy, BLEU score, perplexity (PPL) and POS distance. 
For accuracy, we adopt a CNN-based classifier, trained 
on the same training data, to evaluate whether the generated sentence possesses the 
target style. Then we measure BLEU score and BLEU(human) score of transferred sentences
against the original sentences and ground truth, respectively.  PPL metric is evaluated by GPT-2~\citep{radford2019language} base model after finetuning on the corresponding dataset,
with the goal to assess the fluency of the generated sentence. POS distance is used to measure
the model's semantics preserving ability~\citep{DBLP:journals/corr/abs-1810-06526}. 

We also perform human evaluations on Yelp data to further test the transfer quality. 
We first randomly select 100 sentences from the test set, use these sentences 
as input and generate sentences from the base
model~\citep{DBLP:journals/corr/abs-1810-06526} and our model.
Then for each original sentence, we present the outputs of the base model and ours in random order. 
The three annotators are asked to evaluate which sentence is preferred
as the transferred sentence of the original sentence, in terms of content preservation
and sentiment transfer. They can choose either output or select the same quality.
We measure the percentage of times each model outperforms the other.
\subsubsection{Learning Non-Autoregressive Generation}\label{app:nat_set}
We use the Adam optimizer with learning rate $5\times 10^{-4}$ with inverse square root scheduler. We apply sequence-level knowledge distillation to the dataset, which can reduce the complexity of the dataset, making it easier for the model to learn and improving the performance. The models are first trained by CE loss for fast initialization, then focus on $2$-gram, $3$-gram, and $4$-gram with the same weights.
Fairseq~\citep{fairseq} is adopted to conduct the experiments. 
We average the last 5 checkpoints as the final model.
\subsection{Additional Results of Learning from Noisy Text}\label{app:noisy_results}
\subsubsection{Results of BLEURT Metric}\label{app:noisy_bleurt}
\begin{figure*}[htb]
\centering
\includegraphics[width=\textwidth,page=3]{sections/denoising/exp1results.pdf}
\caption{Results of Translation with Noisy Target on German-to-English(de-en) from Multi30k. BLEURT scores are computed 
against clean test data.}
\label{fig:noisymt_bleurt}
\includegraphics[width=\textwidth,page=1]{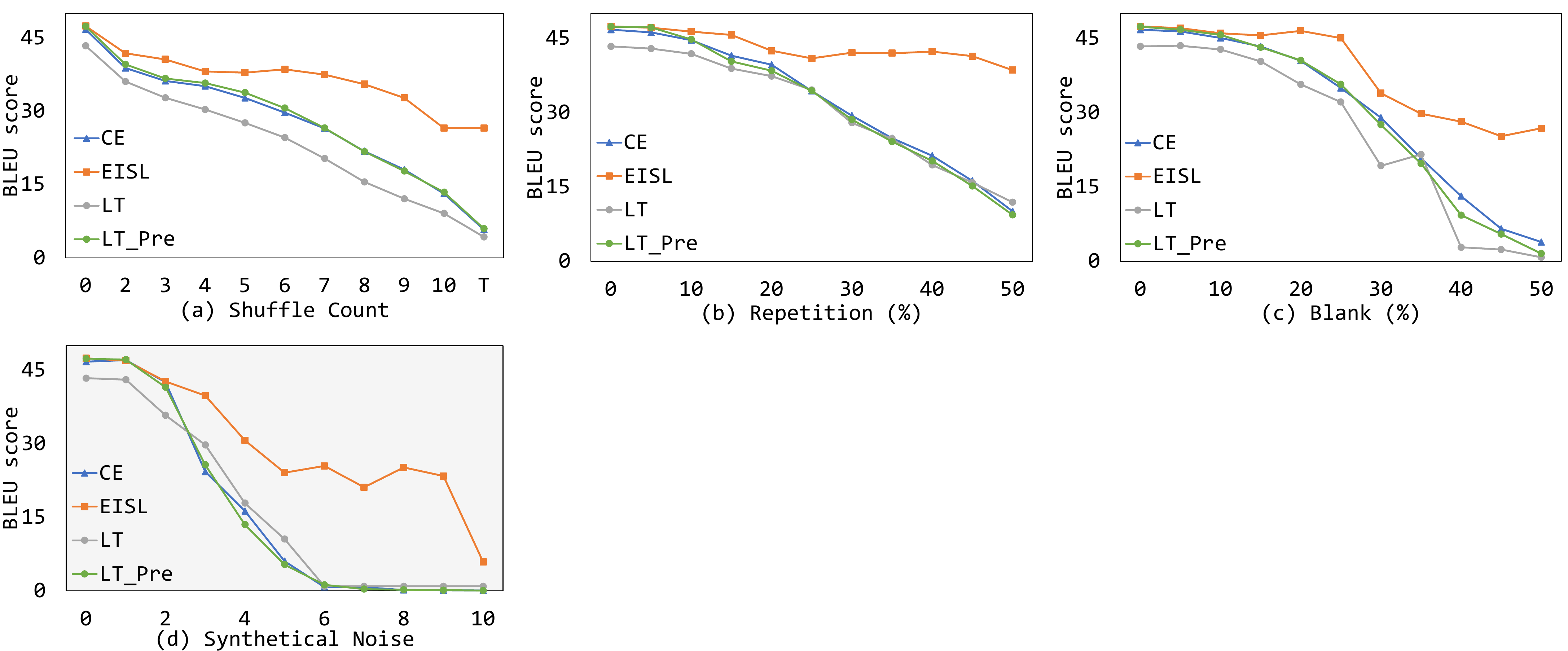}
\vspace{-6pt}
\caption{{Comparison results with Loss Truncation(LT) of Translation with Noisy Target on German-to-English(de-en) from Multi30k. BLEU scores are computed against clean test data.}}
\label{fig:noisy_loss_truncation}
\end{figure*}

In this section, we evaluate the results of CE, PG and EISL on BLEURT~\citep{DBLP:journals/corr/abs-2004-04696} metric. We use the recommended BLEURT-20 checkpoint. It gives a score for every sentence pair, and we averaged the scores to get the final score. The results are shown in Figure~\ref{fig:noisymt_bleurt}. Both BLEU metric and BLEURT metric show the superiority of our proposed EISL loss.

\subsubsection{Comparison with Loss Truncation}
{
The Loss Truncation (LT~\citep{DBLP:conf/acl/KangH20}), method adaptively removes high log loss examples as a way to optimize for distinguishability.
In this section, We'd like to show the comparisons with Loss Truncation.
We evaluated two variants of LT: (1) LT\_Pre which first trains the model with CE loss and then adds LT for further training, and (2) LT which directly trains the model with CE loss and LT together. Hyperparameters were selected on the validation set. For simplicity, we remove the PG curves (Figure~\ref{fig:exp1result}), and the comparison results with LT are shown in Figure~\ref{fig:noisy_loss_truncation}.

We can see Loss Truncation can sometimes slightly improve over CE, especially when the data is clean or with low/moderate noise. However, by simply ignoring high-loss data, LT is not good at handling data with high noise (which often leads to high loss). In comparison, our proposed EISL achieves a substantial improvement in the presence of high noise.
}

\subsubsection{Reasons of Better Performance with Lower-gram EISL}
{
In this section, we discuss the reason of why the performance of using lower grams is better than higher-gram EISL in Figure~\ref{fig:exp1result}(e).
 
Lower-gram EISL is less sensitive to noise. For example, 1-gram EISL focuses mostly on matching individual tokens without caring much about the order of tokens; while a high-gram EISL (e.g., consider the extreme case of 
$T^*$-gram where $T^*$ is the target length) reduces to CE (as discussed in Sec~\ref{sec:connections}) and is highly sensitive to noise. Thus, in the presence of high data noise, lower-gram EISL would be more robust and perform better.

Besides, on low-noise data (e.g., noise-level = 0 or 1), lower-gram EISL performs comparably with higher-gram EISL, both close to the CE performance. This is because we pretrained the model with CE (as mentioned in the experimental setup), and finetuning with EISL (either with lower- or higher-grams) would not change the performance a lot given the low-noise data.
}

\subsubsection{{Cases Study}} \label{app:case_study}
As shown in Table~\ref{example_1}, \ref{example_2}, \ref{example_3}, \ref{example_4} and \ref{example_5}, we randomly sample some examples from generated sentences of the models trained with different types of noise on Multi30k dataset. For the sake of convenience, we use abbreviations in the tables, i.e., SC, RR, BR and NL are short for Shuffle Count, Repetition Ratio, Blank Ratio and Noise Level (for Synthetical Noise), respectively. 

\paragraph{Shuffle Noise} When there exist a few shuffle noises, e.g., SC $=3$, CE loss may lead word reduplicated (Example 1 and Example 2) and slightly wrong word order (Example 4 and Example 5), and there are some information mistranslated (\textit{beautiful} in Example 4) or extra irrelevant information added  (\textit{black} in Example 5). As shuffle count increases, the aforementioned problems are increasingly severe, resulting the generated sentences meaningless. Especially, there are some words untranslated in PG examples (\textit{eingezäunten} in Example 1, \textit{irgendwo} in Example 2, \textit{haben} in Example 5, ). But EISL loss could keep the content consistency and grammatical correctness as far as possible.
\begin{table*}[htb]
\footnotesize
    \begin{center}
        \begin{tabular}{ll}
        \cmidrule[\heavyrulewidth]{1-2}
        Source& my `` hot '' sub was \textit{cold} and the meat was \textit{watery} . \\
        Base Model& my `` hot '' sub was \textit{excellent} and the meat was \textit{excellent} .\\
        \textit{with} EISL&my `` hot '' sub was \textit{delicious} and the meat was \textit{delicious} .\\
        \cmidrule{1-2}
        Source& the man did \textit{not stop} her .\\
        Base Model& the man did \textit{definitely right} her .\\
        \textit{with} EISL&the man did \textit{definitely stop} her .\\
        \cmidrule[\heavyrulewidth]{1-2}
        \end{tabular}
    \caption{Examples of the generated sentences.}
    \label{tab:generatedSent}
    \end{center}
    \begin{center}
    \begin{tabular}{rcccc}
    \cmidrule[\heavyrulewidth]{1-5}
    \multicolumn{1}{r}{\bf Model}  &\multicolumn{1}{l}{\bf Accuracy(\%)}  &\multicolumn{1}{l}{\bf BLEU} &\multicolumn{1}{l}{\bf PPL}
    & \multicolumn{1}{l}{\bf POS distance}
    \\ 
    \cmidrule{1-5}
    \citet{DBLP:conf/acl/TsvetkovBSP18}&86.5 & 7.38 & - & 7.298 \\
      \citet{DBLP:conf/icml/HuYLSX17}&\bf 90.7 & 47.50 & - & 3.524 \\
    \cmidrule{1-5}
    \citet{DBLP:journals/corr/abs-1810-06526} &88.0 & 59.63 & 28.46& 2.348 \\
    \textit{with} \modelname  &{89.2} &\bf{60.26} &\bf{27.85} &\bf{2.191} \\ 
    \cmidrule[\heavyrulewidth]{1-5}
    \end{tabular}
    \vspace{-5pt}
    \caption{The results on the political dataset. The first two results are reported by \cite{DBLP:journals/corr/abs-1810-06526}.}
    \label{tab:politics_result}
    \end{center}
\vspace{-10pt}
\end{table*}
\paragraph{Repetition Noise} The main problem of the models trained by CE and PG with repetition noises is that the models can't filter the repetition noise out in training samples, and try to learn the wrong distribution, leading to generate reduplicated words frequently  (Example 1-5).  Specifically, the examples of CE and PG in RR $=50\%$ are very representative. However, it's amazing that EISL can almost avoid such a problem even the repetition ratio achieves $50\%$. Meanwhile, the main semantics is preserved and the grammar is correct.
\paragraph{Blank Noise} When adding blank noise, some tokens in targets will be substituted as \textit{unk} so the targets will lose some information. We could measure from two aspects: one is the term frequency of meaningless token \textit{unk} in generated sentences, and the other is the meaningful contents preserved by the models. Obviously, EISL loss handles better than CE loss on both aspects. Especially, when BR $=20\%$, unlike models with CE, models with PG and EISL barely generate the \textit{unk} token, and could translate the core content  (Example 1-5). As BR increases, EISL could preserve more key information and produce less \textit{unk} than CE and PG. Moreover, PG performs rather poor when BR is high (like BR $=45\%$), and it almost loses all information (Example 1-5) and generates some confusing words (\textit{teil} in Example 1, \textit{afroamerikanischer} and \textit{irgendwo} in Example 3, \textit{beachaufsichtgebäude} in Example 4, and \textit{holzstück} in Example 5).
\paragraph{Synthetical Noise} We then evaluate the results of models trained by synthetical noise. Such a situation combines aforementioned three types of noises. One most highlighted advantage of EISL is that the generated sentences are almost grammatically correct and include main content as far as possible. However, CE can only stiffly joint some words, and can't guarantee the grammatical correctness (word order, word repetition and so on). PG performs worst, involving all the problems in CE cases and the meaningless word generation problem (Example 1-5).

\subsection{Additional Results of Text Style Transfer}

\subsubsection{Examples on Yelp dataset}
\label{app:tst_yelp}

Some examples of generated sentences are given in Table~\ref{tab:generatedSent}. 
The model with \modelnamett can select more appropriate adjective and improve the quality of the sentences.
In the first example, the model should transfer the negative adjectives \textit{cold} and \textit{watery} to some positive adjectives that describe food. Obviously, the \textit{delicious} is more appropriate than \textit{excellent}. In the second example, the base model reverses both \textit{not} and \textit{stop}, leading to wrong sentiment and inconsistent content. While the model with EISL could avoid such a situation and generate more suitable sentence.
\subsubsection{Results on Political dataset}
\label{app:tst_pol}
Since the instances from democratic data and republican data are quite different, names 
of politicians have high correlation with the political slant. Therefore the BLEU score 
and POS distance have a big gap with the sentiment results. The results are shown in Table~\ref{tab:politics_result}.

\begin{figure*}[htb]
\centering
\includegraphics[width=\linewidth]{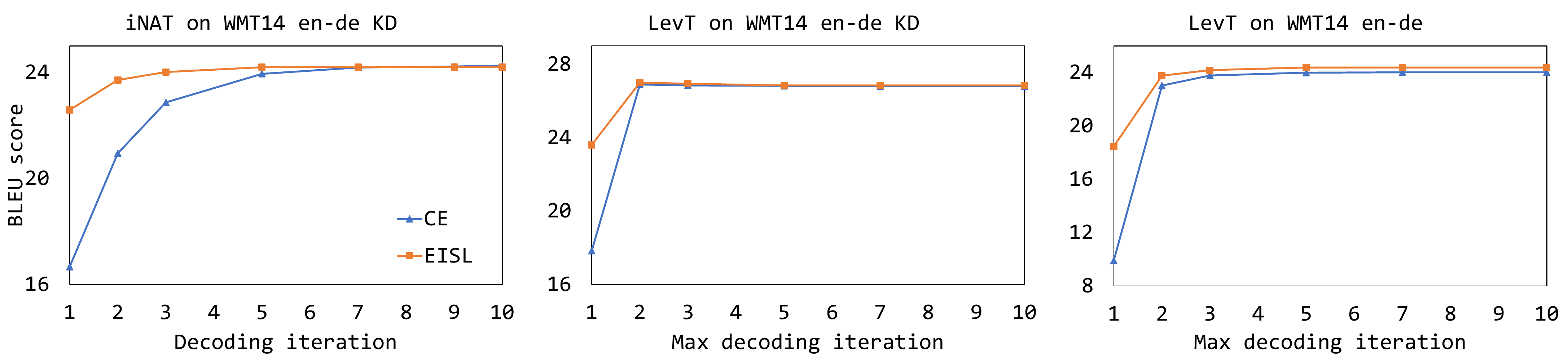}
\caption{Results of iterative NAT on different decoding iterations.}
\label{figure:nat-iter}
\end{figure*}
\subsection{Additional Results of Non-Autoregressive Generation}





\subsubsection{Results of Iterative NAT Models}
\label{app:nat_iter}
As shown in Figure \ref{figure:nat-iter}, with the increasing of iteration steps, the difference fades away. 

\subsubsection{Results of BLEURT Metric}\label{app:nat_bleurt}
To show the superiority of our method, We also evaluate on recent text generation metric, BLEURT~\citep{DBLP:journals/corr/abs-2004-04696}. BLEURT is an evaluation metric for Natural Language Generation. It takes a pair of sentences as input, a reference and a candidate, and it returns a score that indicates to what extent the candidate is fluent and conveys the mearning of the reference. We use the recommended BLEURT-20 checkpoint. It gives a score for every sentence pair, and we averaged the scores to get the final score. The results are shown in Table~\ref{table:NAT_BLEURT}. 
\begin{table*}[htb]
\footnotesize
  \centering
  \begin{tabular}{lcccc}
    \cmidrule[\heavyrulewidth]{1-5}
      \multirow{2}{*}{\bf Model}  
     &  \multicolumn{2}{c}{\bf WMT14 en-de KD}   &   \multicolumn{2}{c}{\bf WMT14 en-de} \\
    \cmidrule(lr){2-3}\cmidrule(lr){4-5} 
      & CE & \modelname & CE & \modelname \\
    \midrule
  Vanilla-NAT~\citep{natbase}  &   0.346   &   {\bf0.416} &   0.194  & {\bf0.277}   \\
    NAT-CRF~\citep{DCRF} & 0.441 & \bf 0.464 & -& - \\
    iNAT~\citep{inat}   &0.332  &  {\bf0.437}  & - & - \\
    LevT~\citep{levt}  & 0.355& \bf 0.458& 0.214 &\bf 0.333 \\
    CMLM~\citep{cmlm} & 0.345	& \bf 0.450  & - &  - \\
    \cmidrule[\heavyrulewidth]{1-5}
  \end{tabular}
    \caption{
    The results (test set BLEURT) of \modelnamett loss and CE loss applied to non-autoregressive models.}
    \label{table:NAT_BLEURT}
\end{table*}

\subsubsection{Qualitative Analysis on NAT Experiments}
\label{app:nat_qual}

Given the non-autoregressive nature (i.e., all tokens are generated simultaneously), the one-to-one matching of CE loss can lead to severe mismatching. We consider the example: the predicted sentence is \texttt{a cat is on the red blanket} and the target sentence is \texttt{a cat is sitting on the red blanket}. The "on the red blanket" part of the prediction will be corrected to match the target positions, and this may lead to overcorrection (e.g., "on the red red blanket ."). Repetition is often a sign of overcorrection. However, with EISL, this situation will not happen because the phrase will be matched to appropriate target tokens. Let's have a look at a real example in 
Figure~\ref{fig:nat_analysis}.
\begin{figure}[htb]
    \begin{center}
    \footnotesize
        \begin{tabular}{ll}
        \cmidrule[\heavyrulewidth]{1-2}
        Source& Anja Schlichter managed the tournament \\
        Target& Anja Schlichter leitet das Turnier\\
        CE & Anja Schlichter leitdas Turnier Turnier\\
        EISL & Anja Schlichter leitete das Turnier geleitet\\
        \cmidrule[\heavyrulewidth]{1-2}
        \end{tabular}
    \caption{Examples of the generated sentences.}
    \label{fig:nat_analysis}
    \end{center}
\end{figure}

\begin{figure}[htb]
\centering
\includegraphics[width=0.45\textwidth,page=2]{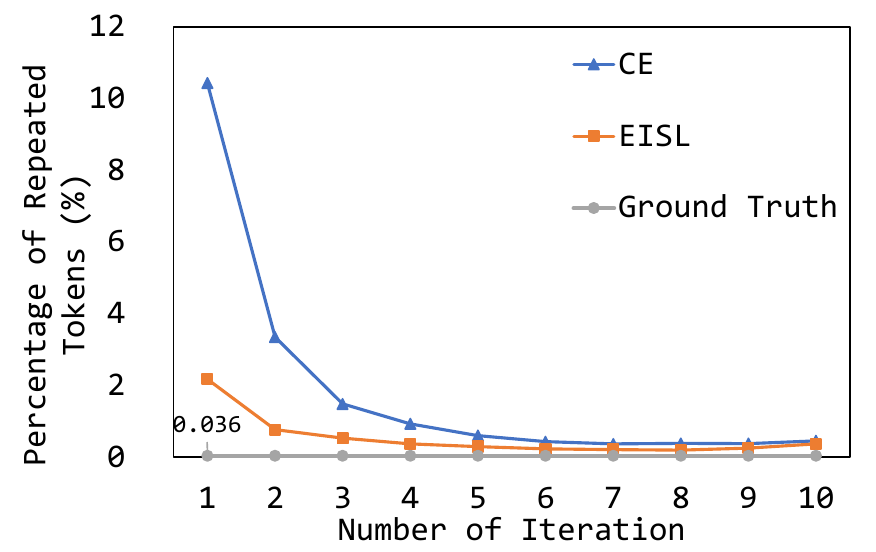}
\caption{The percentage of repeated tokens under different iteration steps.}
\label{fig:nat_repeat}
\end{figure}

Take the non-autoregressive model CMLM~\citep{cmlm} for example, we evaluate the translation of CMLM models trained by CE and EISL. As shown in Figure \ref{fig:nat_repeat}, our proposed \modelnamett can reduce repetition to a large extent.

\subsection{{Efficiency Analysis}}

\paragraph{Complexity analysis}
Given $T^*$ tokens, the time complexity of CE loss is $\mathbf O(T^*)$, while the complexity of $n$-gram EISL loss is $\mathbf O(n(T^*-n+1)^2)\approx\mathbf O({T^*}^2)$,  assuming small $n$ is used in practice (e.g., $n\in \{1,2,3,4\}$). However, in practice, the computation cost of the loss (either CE or EISL) is {\bf negligible} compared to the cost of model forward and backward during training. Thus, the extra cost introduced by the EISL loss is rather minor.

\begin{figure*}[htb]
\centering
\includegraphics[width=0.95\textwidth,page=2]{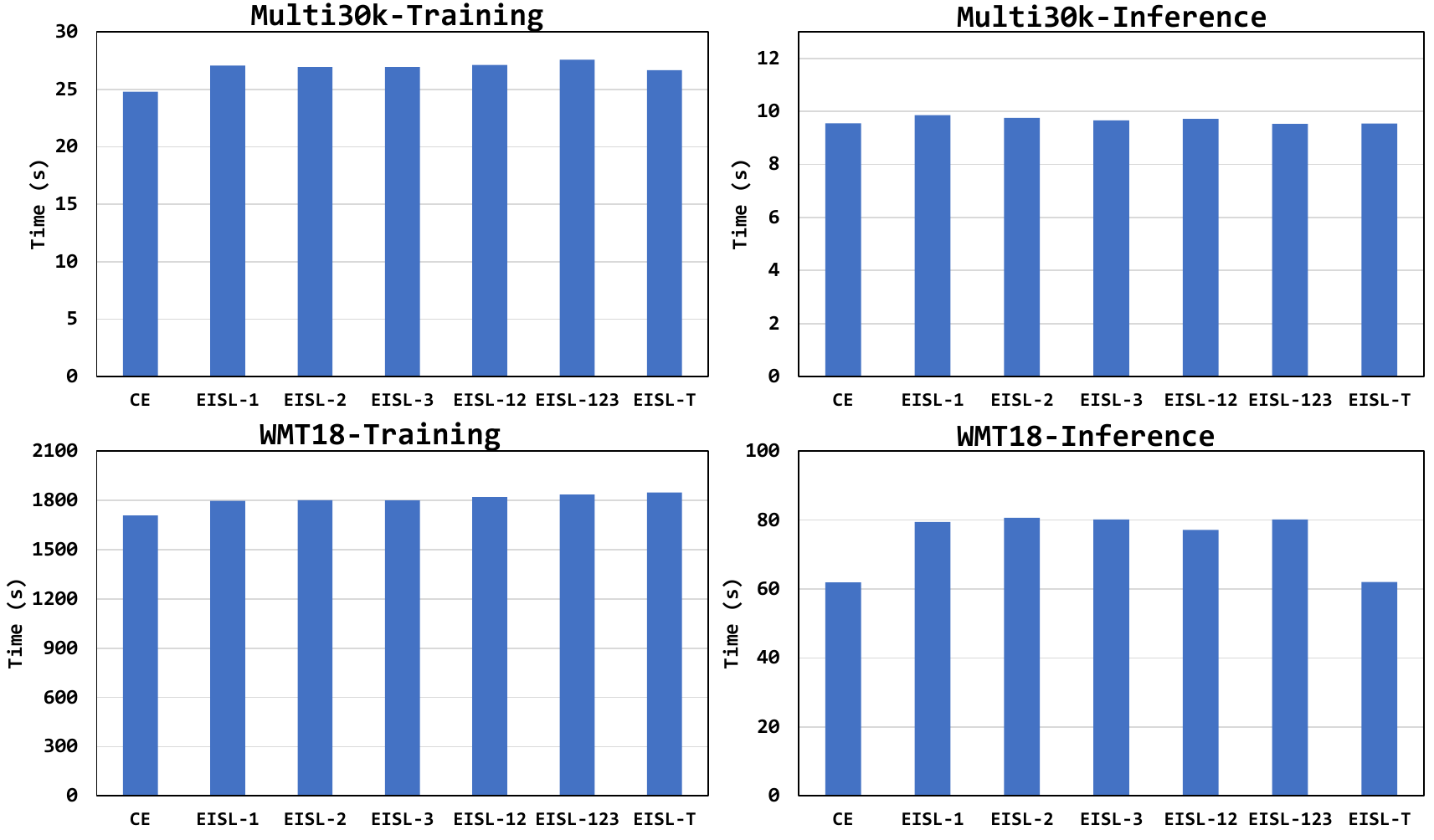}
\vspace{-10pt}
\caption{Results of training and inference time. EISL-$n$ represents $n$-gram EISL loss and EISL-$12$ represents the combination of 1-gram and 2-gram EISL loss.}
\label{fig:complexity}
\end{figure*}

\paragraph{Empirical comparison of time cost}
To quantify the computational cost of different methods, we adopt CE and EISL on top of the same model and setting, and evaluate the consumed time for 1 training epoch. For comparison on both small and large dataset, we evaluate on Multi30k (29k training data, 1k test data) and 1M scale WMT-18 raw corpus (1M training data, 3k test data). The models are tested on one Tesla V100 DGXS with 32 GB memory, the batch size is 128, max number of tokens is 6000 and update frequency is 4. For each method, we test 6 times and average the results as final time. The results are shown in Figure~\ref{fig:complexity}.

\paragraph{Empirical total time cost of EISL training} 
As discussed in the experiments in the paper, we first pretrain the model with the CE loss until convergence, and then finetune with the EISL loss. Here we report the total time cost of each stage, based on the WMT-18 translation setting as described in Section~\ref{sec:noisy}. The results are shown in Table~\ref{tab:empirical_time}. As the data size increases, the convergence time of both pretraining and finetuning grows. The time cost of the finetuning stage is less than half of that of the pretraining stage.
\begin{table*}[htb]
    \begin{center}
    \footnotesize
        \begin{tabular}{crr}
        \cmidrule[\heavyrulewidth]{1-3}
        Data Size & PreTraining Time (CE) & Finetuning Time (EISL) \\ 
        \cmidrule[\heavyrulewidth]{1-3}
        1M & 1h 40min 57s&49min 33s \\ 
        2M& 5h 56min 57s&1h 35min 10s \\ 
        4M& 8h 55min 18s&3h 57min 44s\\ 
        \cmidrule[\heavyrulewidth]{1-3}
        \end{tabular}
    \caption{
    Convergence time of pretraining and finetuning stages. 
    }
    \label{tab:empirical_time}
    \end{center}
\end{table*}

\subsection{{Hyperparameters}}
Regarding which $n$-grams to use and their weights $w_n$ in the EISL loss, we found in our experiments that the default values \emph{largely} following the standard BLEU metric (i.e., maximum $n=4$ with equal weights) work well. Specifically, we use $n\in\{2,3,4\}$ and equal weights $w_n=1/3$ as our default values. Most of our experiments adopt the default values which achieve consistent substantial improvement over CE and other rich baselines as shown in our experiments. (except for the synthetic experiment where we show the effect of different $n$-grams including those selected using the validation set).

Besides, in our experiments, we first pretrain the model with the CE loss (i.e., EISL with $n=T^*$ and teacher forcing, see Section~\ref{sec:connections}) and then finetune with the EISL loss. We simply do the CE pretraining \emph{until convergence} before switching to the EISL finetuning. Therefore, there is no need of tuning for the training iterations of pretraining.

\subsection{{Analysis of Efficient Implementation}}\label{sec:app:approximation}
\begin{figure*}[htb]
\centering
\includegraphics[width=\textwidth]{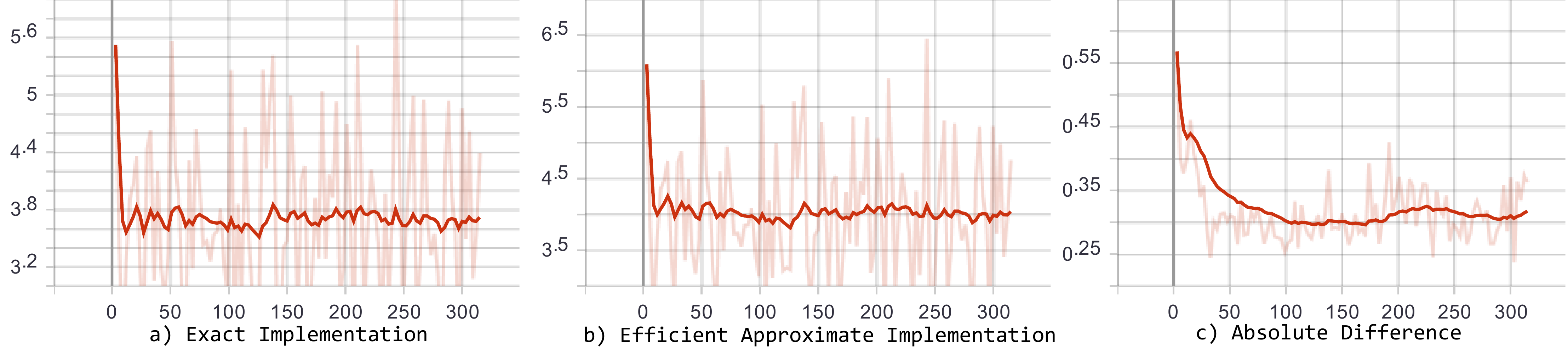}
\vspace{-15pt}
\caption{The change of loss values during training. The x-axis represents the training step. a) gives the loss curve of exact implementation; b) gives the loss curve of efficient approximate implementation as we discussed in section~\ref{para:approx_p}; and c) gives the absolute difference between the two implementations.}
\label{fig:approx_p}
\end{figure*}

In order to validate the efficiency and accuracy of our approximation (for autoregressive models) discussed in Section~\ref{para:approx_p}, we conduct the analysis experiments, showing that the approximate (and efficient) EISL loss values are very close to exact (but expensive) EISL value. We use the same setting as section~\ref{sec:noisy}, and finetune the model with our efficient approximate EISL loss on Multi30k. Throughout the course of training, we record the loss values of both the exact implementation and our approximate implementation.
As shown in Figure~\ref{fig:approx_p}(a) and (b), the tendency of two losses is very close to each other. We also plot the absolute difference of the two losses as shown in Figure~\ref{fig:approx_p}(c). We can see the difference decreases as training proceeds. The observations validate the effectiveness of our approximate implementation. 

We note that training the model with the exact loss is costly, which necessitates our approximation. Specifically, for $n$-gram loss, we need to run the forward pass of the decoder $(T-n)^2$ times, and keep the whole computation graph for back-propagation, which will consume much more time and memory. Even for only loss evaluation (without the backward pass), we found the runtime of the exact loss is about 15 times longer than that of the efficient approximate implementation based on convolution operator.

\begin{table*}[htb]
    \begin{center}
    \footnotesize
        \begin{tabular}{cll}
        \cmidrule[\heavyrulewidth]{1-3} 
        \multicolumn{2}{c}{Source (de)}& \makecell[l]{ein junger mann nimmt an einem lauf teil und derjenige , der dies aufzeichnet ,\\ lächelt .}\\
        \midrule
        \multicolumn{2}{c}{Target (en)}&\textbf{a young man participates in a career while the subject who records it smiles .}\\
         \cmidrule[\heavyrulewidth]{1-3} 
        \multirow{3}{*}{SC $=3$}&CE&  young man is running on a a and the other man is smiling .\\
        &PG&  young man is running on a track and the other man is smiling .\\
        &EISL& young man is running in a dirt course and the other is smiling . \\
        \midrule
        \multirow{3}{*}{SC $=6$}&CE&  young man is running a a race and the other is smiling .\\
        &PG&  young man taking a race and the other smiling . a\\
        &EISL& young man is running a race and the other guy is smiling .\\
        \midrule
        \multirow{3}{*}{SC $=9$}&CE& young man . a a the is running up and up hill smiling taking\\
        &PG&  young man takes on a slope and thejenige , the the smiles . a\\
        &EISL& young man is on a hillside smiling and the others , who is smiling .\\
        \cmidrule[\heavyrulewidth]{1-3} 

        \multirow{3}{*}{RR $=15\%$}&CE&  young man is running on a track and the other is smiling .\\
        &PG& young man is running on a track and the other is smiling . \\
        &EISL&  young man is running in a race and the runner is smiling .\\
        \midrule
        \multirow{3}{*}{RR $=30\%$}&CE&  \makecell[l]{young man man is is running on a track track and the the other is is smiling\\ smiling .}\\
        &PG&   \makecell[l]{young man man is is running on a track track and the other man man who is \\is is smiling .}\\
        &EISL& young man is running in a race and the other is smiling at him . .\\
        \midrule
        \multirow{3}{*}{RR $=50\%$}&CE&  \makecell[l]{a young young man man is is smiling smiling at at a a window window while\\ another smiles smiles at him him . .}\\
        &PG&   \makecell[l]{a young man man is is napping napping on on a a grassy grassy field field and\\ and some people people are are smiling smiling . .}\\
        &EISL&  young man running in a race and the other is smiling at the action . .\\
        
        \cmidrule[\heavyrulewidth]{1-3} 
        \multirow{3}{*}{BR $=20\%$}&CE&  young man unk unk a run and the unk is smiling .\\
        &PG&  young man is running in a race and the one who is looking at him is smiling .\\
        &EISL& young man is running in a race with the runner who is up .\\
        \midrule
        \multirow{3}{*}{BR $=35\%$}&CE& young man unk unk a unk , and the unk is smiling unk\\
        &PG& young man unk unk track unk others unk . \\
        &EISL&  young man unk is un in a race and the other un is un at the finish .\\
        \midrule
        \multirow{3}{*}{BR $=45\%$}&CE&  young unk is unk on a unk unk and the unk smiles unk\\
        &PG&  young man unk a unk teil unk unk .\\
        &EISL&  young unk un is un in a race , the other is smiling back .\\
        
        \cmidrule[\heavyrulewidth]{1-3} 
        \multirow{3}{*}{NL $=5$}&CE&  young man is running a race and the one who is running is smiling .\\
        &PG&  young man is running a race and the one scoring is smiling .\\
        &EISL&  young man is running a race and one of the runners is up to him .\\
        \midrule
        \multirow{3}{*}{NL $=15$}&CE&  young man is unk unk a unk and the other man is smiling . \\
        &PG& young man is on a unk smiling at thejenige . .\\
        &EISL&  young man is in a race , the other smiling . \\
        \midrule
        \multirow{3}{*}{NL $=20$}&CE&   a young man is unk unk a unk and unk is smiling at him . \\
        &PG& young smiles on in ail and thejenige smile on . . . \\
        &EISL&   young man unk unk a ladder and unk , who is unk smiling . \\
        \cmidrule[\heavyrulewidth]{1-3} 
        \end{tabular}
    \caption{Example 1. }
    \label{example_1}
    \end{center}
\end{table*}

\begin{table*}[htb]
    \begin{center}
    \footnotesize
        \begin{tabular}{cll}
        \cmidrule[\heavyrulewidth]{1-3} 
        \multicolumn{2}{c}{Source (de)}&15 große hunde spielen auf einem eingezäunten grundstück neben einem haus .\\
        \midrule
        \multicolumn{2}{c}{Target (en)}&\textbf{15 large dogs playing in a fenced yard beside a house .}\\
         \cmidrule[\heavyrulewidth]{1-3} 
        \multirow{3}{*}{SC $=3$}&CE& large dogs play on a a dirt path next to a house .\\
        &PG& 15 large dogs play on an earthen platform next to a house .\\
        &EISL&  large dogs are playing on a dirt path next to a house .\\
        \midrule
        \multirow{3}{*}{SC $=6$}&CE&  large dogs play on a a play area next to abandoned house .\\
        &PG& 15 large dogs playing on a eingezäunten group stage next to a house .\\
        &EISL&  group of dogs play on a abandoned path next to a house .\\
        \midrule
        \multirow{3}{*}{SC $=9$}&CE& large dogs play a . on a field next to abandoned house\\
        &PG&  dogs play on a snowy grundstück next to a house .15 large\\
        &EISL&  . 15 large dogs play on an abandoned hillside next to a house .\\
        \cmidrule[\heavyrulewidth]{1-3} 

        \multirow{3}{*}{RR $=15\%$}&CE&   large dogs are playing on a fenced in area next to a house .\\
        &PG&  large dogs are playing on a fenced in area next to a house .\\
        &EISL&  large dogs are playing on a fenced track next to a house .\\
        \midrule
        \multirow{3}{*}{RR $=30\%$}&CE&  large dogs dogs play on on a a dirt track near a house house .\\
        &PG&  large dogs dogs play on a fenced-in area area next to a house .\\
        &EISL& large dogs play on a fenced walkway next to a house . .\\
        \midrule
        \multirow{3}{*}{RR $=50\%$}&CE&\makecell[l]{small dogs dogs play on on a a grassy grassy field field next next to to a house \\house . .}\\
        &PG& \makecell[l]{15 large dogs dogs are are playing playing on on a a grassy grassy field field\\ next next to to a house house . .}\\
        &EISL&15 large dogs playing on a fenced terrain next to a house . .\\
        
        \cmidrule[\heavyrulewidth]{1-3} 
        \multirow{3}{*}{BR $=20\%$}&CE& large dogs play in a fenced yard next to a house .\\
        &PG&  large dogs are playing on an overcast walk next to a house .\\
        &EISL&  large dogs are playing in a fenced area near to a house .\\
        \midrule
        \multirow{3}{*}{BR $=35\%$}&CE& unk dogs play unk a unk unk by a house .\\
        &PG&  large dogs unk a unk path unk unk house .\\
        &EISL&   large dogs unk play in a fenced area next to a house .\\
        \midrule
        \multirow{3}{*}{BR $=45\%$}&CE&  unk dogs unk on a unk unk next to unk house .\\
        &PG&  large dogs unk a unk unk .\\
        &EISL&   large unk un are un in a fenced-out game next to a house .\\
        
        \cmidrule[\heavyrulewidth]{1-3} 
        \multirow{3}{*}{NL $=5$}&CE&  large dogs are playing on a fenced in area next to a house .\\
        &PG& large dogs are playing on a fenced in area next to a house .\\
        &EISL&   large dogs are playing on a fenced backwalk next to a house . \\
        \midrule
        \multirow{3}{*}{NL $=15$}&CE&  large dogs are playing on a unk grassy field next to a house .\\
        &PG&  large dogs playing on a unk next to a house . . .\\
        &EISL&   large dogs play on a covered piece of furniture next to a house .\\
        \midrule
        \multirow{3}{*}{NL $=20$}&CE&  large dogs are playing on on a a a grassy grassy field next to a house .\\
        &PG& large play play in auntenck in a house . . . \\
        &EISL&  large dogs play on a unk unk next to a house . . \\
        \cmidrule[\heavyrulewidth]{1-3} 
        \end{tabular}
    \caption{Example 2.}
    \label{example_2}
    \end{center}
\end{table*}

\begin{table*}[htb]
    \begin{center}
    \footnotesize
        \begin{tabular}{cll}
        \cmidrule[\heavyrulewidth]{1-3} 
        \multicolumn{2}{c}{Source (de)}&ein afroamerikanischer mann spielt irgendwo in der stadt gitarre und singt\\
        \midrule
        \multicolumn{2}{c}{Target (en)}&\textbf{an african american man playing guitar and singing in an urban setting .}\\
         \cmidrule[\heavyrulewidth]{1-3} 
        \multirow{3}{*}{SC $=3$}&CE& african american man is playing the guitar and singing in the city .\\
        &PG&  african american man is playing the guitar in the city and singing\\
        &EISL&    african american man is playing the guitar in the city and singing .\\
        \midrule
        \multirow{3}{*}{SC $=6$}&CE&  african-american man is playing guitar in the a and singing city .\\
        &PG& african american man playing irgendwo in the city guitar singing\\
        &EISL& african american man is playing the guitar in the city\\
        \midrule
        \multirow{3}{*}{SC $=9$}&CE&  african-american man playing guitar in the a and singing city\\
        &PG& african americanischer man plays irgendwo in the city guitar singing . a
        \\&EISL&  african american man is playing the guitar in the city and singing\\
        \cmidrule[\heavyrulewidth]{1-3} 
        \multirow{3}{*}{RR $=15\%$}&CE&  african american american man plays guitar guitar in the city city .\\
        &PG&  african american man is playing guitar in the city and singing .\\
        &EISL&  african american man is playing guitar in the city and singing .\\
        \midrule
        \multirow{3}{*}{RR $=30\%$}&CE&   african american man plays guitar guitar in in the city city while singing .\\
        &PG&  african american man man plays guitar guitar in the city city and sings .\\
        &EISL& an african american man playing guitar in the city and singing . .\\
        \midrule
        \multirow{3}{*}{RR $=50\%$}&CE&\makecell[l]{ african african american american man playing guitar guitar in in the the\\ city city and singing singing .}\\
        &PG& \makecell[l]{african american american man man is is playing playing guitar guitar\\ in in the the city city . .}\\
        &EISL&  an african american man playing guitar in the city and singing . .\\
        \cmidrule[\heavyrulewidth]{1-3} 
        \multirow{3}{*}{BR $=20\%$}&CE&  african american man plays guitar unk sings unk\\
        &PG&  african american man is playing guitar and singing in the city .\\
        &EISL&   african american man is playing the guitar and singing .\\
        \midrule
        \multirow{3}{*}{BR $=35\%$}&CE&  african american man unk unk guitar unk singing unk\\
        &PG&  african american man unk guitar unk singing unk\\
        &EISL&    african american unk is un a guitar and singing in the city .\\
        \midrule
        \multirow{3}{*}{BR $=45\%$}&CE&   african american unk unk playing unk guitar in unk city unk\\
        &PG&  afroamerikanischer man unk irgendwo unk unk\\
        &EISL&    af unk un playing some sort of guitar in the city and singing .\\
        
        \cmidrule[\heavyrulewidth]{1-3} 
        \multirow{3}{*}{NL $=5$}&CE&  african american man plays guitar and sings somewhere in the city . \\
        &PG&  african american man is playing guitar and singing in the city .\\
        &EISL&  african american man is playing guitar and singing somewhere in the city . \\
        \midrule
        \multirow{3}{*}{NL $=15$}&CE&  african american man is playing the guitar in the city and singing .\\
        &PG&  afroamerikanischer man is irgendwo in the city guitarre .\\
        &EISL& african american man playing some sort of guitar in the city and singing .  \\
        \midrule
        \multirow{3}{*}{NL $=20$}&CE&   african american american man is playing the guitar in the the city unk\\
        &PG&  afroamerikanischer singt in the city guitarre singt .\\
        &EISL&  african american man plays unk unk in the city unk\\
        \cmidrule[\heavyrulewidth]{1-3} 
        \end{tabular}
    \caption{Example 3.}
    \label{example_3}
    \end{center}
\end{table*}

\begin{table*}[htb]
    \begin{center}
    \footnotesize
        \begin{tabular}{cll}
        \cmidrule[\heavyrulewidth]{1-3} 
        \multicolumn{2}{c}{Source (de)}&ein strandaufsichtgebäude steht im sand , es ist ein bewölkter tag .\\
        \midrule
        \multicolumn{2}{c}{Target (en)}&\textbf{a lifeguard building is on the sand on a cloudy day .}\\
         \cmidrule[\heavyrulewidth]{1-3} 
        \multirow{3}{*}{SC $=3$}&CE&  beach a is standing in the sand on a beautiful day .\\
        &PG& beachfront building is standing in the sand on a beautiful day .\\
        &EISL&   beach view building is standing in the sand on a cloudy day .\\
        \midrule
        \multirow{3}{*}{SC $=6$}&CE&   beach a is in the sand building on a beautiful day .\\
        &PG& beach viewgeb building standing in sand on a beautiful day .\\
        &EISL&  beach view building is standing in the sand on a beautiful day . \\
        \midrule
        \multirow{3}{*}{SC $=9$}&CE&  beach a in the sand . a cloudy day stands beach\\
        &PG&  beachaufsichtge building stands in sand , the is a beautiful day . a\\
        &EISL&   . a beachfront building standing in the sand is a beautiful day .\\
        \cmidrule[\heavyrulewidth]{1-3} 
        \multirow{3}{*}{RR $=15\%$}&CE&   beachfront building is standing in the sand on a cloudy day .\\
        &PG&  beachfront building is standing in sand , it is a cloudy day .\\
        &EISL&  beach building is standing in the sand , it is a cloudy day .\\
        \midrule
        \multirow{3}{*}{RR $=30\%$}&CE&  \makecell[l]{beachfront beachfront building building is is standing standing in the sand \\sand on a cloudy day .}\\
        &PG& \makecell[l]{ beachfront beachfront building building is standing in sand sand on a cloudy\\ day .}\\
        &EISL&  beachfront building is standing in the sand , it is a cloudy day . .\\
        \midrule
        \multirow{3}{*}{RR $=50\%$}&CE&\makecell[l]{ a beachfront beachfront building building is is standing standing in in the\\ sand sand , it looks like it is is a beach resort resort . .}\\
        &PG& \makecell[l]{a beachfront beachfront building building is is standing standing in in sand\\ sand . .}\\
        &EISL&   a beach view building is in the sand , it is a cloudy day . .\\
        \cmidrule[\heavyrulewidth]{1-3} 
        \multirow{3}{*}{BR $=20\%$}&CE&  beachfront building is standing in sand on a cloudy day unk\\
        &PG&  beachfront building is standing in sand on a cloudy day .\\
        &EISL&    beach view building is standing in the sand , it is a cloudy day .\\
        \midrule
        \multirow{3}{*}{BR $=35\%$}&CE&   beach unk unk standing in sand on a cloudy day unk\\
        &PG&  beach unk building unk unk sand unk a cloudy day .\\
        &EISL&    beach building unk is un in the sand on a cloudy day .\\
        \midrule
        \multirow{3}{*}{BR $=45\%$}&CE&  unk unk is standing unk the sand unk it is a beautiful day unk\\
        &PG&  beachaufsichtgebäude unk unk sand unk .\\
        &EISL&     beach unk un is un in the sand , this is a cloudy day .\\
        
        \cmidrule[\heavyrulewidth]{1-3} 
        \multirow{3}{*}{NL $=5$}&CE&  beachfront view building is standing in the sand on a cloudy day . \\
        &PG&   beachfront view building is standing in sand on a cloudy day .\\
        &EISL&   beachfront building is standing in the sand , it is a cloudy day . \\
        \midrule
        \multirow{3}{*}{NL $=15$}&CE&  beach unk unk is standing in the sand unk it is a sunny day . \\
        &PG&  beach unk is in sand on a snowy day . . \\
        &EISL&   beach building is in the sand , it is a cloudy day . \\
        \midrule
        \multirow{3}{*}{NL $=20$}&CE&   beach unk unk is standing in the sand unk it is a sunny sunny day . \\
        &PG&  beachaufsichtgebäude steht in sand , es is a day . .\\
        &EISL&   beach unk stands in sand unk it is a sunny day . .\\
        \cmidrule[\heavyrulewidth]{1-3} 
        \end{tabular}
    \caption{Example 4.}
    \label{example_4}
    \end{center}
\end{table*}

\begin{table*}[htb]
    \begin{center}
    \footnotesize
        \begin{tabular}{cll}
        \cmidrule[\heavyrulewidth]{1-3} 
        \multicolumn{2}{c}{Source (de)}&zwei hunde haben beim spielen dasselbe holzstück im maul .\\
        \midrule
        \multicolumn{2}{c}{Target (en)}&\textbf{two dog is playing with a same chump on their mouth .}\\
         \cmidrule[\heavyrulewidth]{1-3} 
        \multirow{3}{*}{SC $=3$}&CE&  dogs are two playing with . pieces of wood in their mouths two\\
        &PG&  dogs are playing with pieces of black wood in their mouths .\\
        &EISL&  two dogs are playing with pieces of wood in their mouths .\\
        \midrule
        \multirow{3}{*}{SC $=6$}&CE&  dogs are two . playing with sticks in their mouths two\\
        &PG&  dogs have been playing with pieces of wood in their mouths . two\\
        &EISL&   two dogs are playing with pieces of wood in their mouths .\\
        \midrule
        \multirow{3}{*}{SC $=9$}&CE&   two dogs their . are playing with sticks in muzzled\\
        &PG& dogs haben beim play pieces in their mouth . two\\
        &EISL&    . two dogs have been playing with sticks in their mouth .\\
        \cmidrule[\heavyrulewidth]{1-3} 
        \multirow{3}{*}{RR $=15\%$}&CE&  two dogs are are playing with a a piece piece of wood in their mouth .\\
        &PG&  dogs are playing with white wooden blocks in their mouth .\\
        &EISL&  two dogs are playing with some pieces of wood in their mouths .\\
        \midrule
        \multirow{3}{*}{RR $=30\%$}&CE&  \makecell[l]{ two dogs dogs are are playing with a a piece piece of of wood in their mouths .}\\
        &PG&  dogs dogs are are playing with white wooden blocks blocks in their mouth .\\
        &EISL&  two dogs are playing with pieces of wood in their mouths . .\\
        \midrule
        \multirow{3}{*}{RR $=50\%$}&CE&\makecell[l]{ two dogs dogs are are playing playing with with plastic plastic sticks sticks in\\ in their their mouth mouth . .}\\
        &PG&  \makecell[l]{two dogs dogs are are playing playing with with plastic holsters holsters in in\\ their maul maul . .}\\
        &EISL&  two dogs have playing with some white wood in their mouths . .\\
        \cmidrule[\heavyrulewidth]{1-3} 
        \multirow{3}{*}{BR $=20\%$}&CE&  dogs unk unk pieces of wood in their mouths .\\
        &PG&  dogs are playing with wet wood in their mouths .\\
        &EISL&    dogs are playing with wet pieces of wood in their mouths .\\
        \midrule
        \multirow{3}{*}{BR $=35\%$}&CE&  unk have unk pieces of unk in their mouths .\\
        &PG&  two dogs unk unk piece of wood unk their mouth .\\
        &EISL&   two dogs unk playing with some piece of wood in their mouth .\\
        \midrule
        \multirow{3}{*}{BR $=45\%$}&CE&  dogs are playing with unk unk in unk mouth unk\\
        &PG&  dogs unk unk piece of unk holzstück unk .\\
        &EISL&    dogs unk un are un while play with some wood pieces in their mouth .\\
        
        \cmidrule[\heavyrulewidth]{1-3} 
        \multirow{3}{*}{NL $=5$}&CE&  two dogs are playing with the same piece of wood in their mouths .\\
        &PG&   dogs have pieces of of wood in their mouths .\\
        &EISL&  two dogs are playing with the same piece of wood in their mouths .\\
        \midrule
        \multirow{3}{*}{NL $=15$}&CE&  two dogs are are are playing with unk unk in their mouths . \\
        &PG&  dogs haben on a game unk unk . . .\\
        &EISL&  two dogs have been playing with a piece of wood in their mouth . \\
        \midrule
        \multirow{3}{*}{NL $=20$}&CE&   two dogs are are are playing with unk unk in their mouths . \\
        &PG&  dogs haben in a playenselbeck in their mouth . .\\
        &EISL&   two dogs are playing with unk sticks in their mouths . . \\
        \cmidrule[\heavyrulewidth]{1-3} 
        \end{tabular}
    \caption{Example 5.}
    \label{example_5}
    \end{center}
\end{table*}


\end{document}